\def\cleanpreview{1}
\documentclass{article}

\usepackage{iclr2027_conference}

\usepackage{newtxtext,newtxmath}
\usepackage{microtype}
\makeatletter
\renewenvironment{abstract}{\vskip.075in\centerline{\large\bfseries
Abstract}\vspace{0.5ex}\begin{quote}}{\par\end{quote}\vskip 1ex}
\def\section{\@startsection{section}{1}{\z@}{-2.0ex plus
    -0.5ex minus -.2ex}{1.5ex plus 0.3ex minus 0.2ex}{\large\bfseries\raggedright}}
\def\subsection{\@startsection{subsection}{2}{\z@}{-1.8ex plus
    -0.5ex minus -.2ex}{0.8ex plus .2ex}{\normalsize\bfseries\raggedright}}
\def\subsubsection{\@startsection{subsubsection}{3}{\z@}{-1.5ex plus
    -0.5ex minus -.2ex}{0.5ex plus .2ex}{\normalsize\bfseries\raggedright}}
\makeatother
\usepackage{enumitem}
\usepackage{algorithm}
\usepackage{algpseudocode}

\usepackage{amsmath,amsfonts,bm}

\def\eqref#1{equation~\ref{#1}}

\def\1{\bm{1}}

\DeclareMathAlphabet{\mathsfit}{\encodingdefault}{\sfdefault}{m}{sl}
\SetMathAlphabet{\mathsfit}{bold}{\encodingdefault}{\sfdefault}{bx}{n}

\usepackage{wrapfig}
\usepackage{booktabs}
\usepackage{placeins}
\usepackage[font=small,labelfont=bf,labelsep=period,
            justification=raggedright,singlelinecheck=false]{caption}
\usepackage{graphicx}
\usepackage{multirow}
\usepackage{array}
\usepackage{tcolorbox}
\tcbuselibrary{breakable}
\usepackage{wrapfig}
\usepackage{xspace}
\usepackage{hyperref}
\usepackage{url}
\hypersetup{hidelinks}

\newcommand{\method}{\textsc{SkillScaffold}\xspace}
\newcommand{\ours}{\textbf{\method}}

\title{SkillScaffold: Scaffolding Agentic Reinforcement Learning with Policy-Adaptive Skill Credit Assignment and Selective Injection}

\author{
  \textbf{Wenlin Zhong\textsuperscript{1,3}, Hao Wang\textsuperscript{2,3}, Kun Yuan\textsuperscript{3}\thanks{Project leader.}, Fan Yang\textsuperscript{3}, Jinluan Yang\textsuperscript{1}}\\
  \textbf{Dian Jin\textsuperscript{1}, Xu Tan\textsuperscript{4}, Yiquan Wu\textsuperscript{1}\thanks{Corresponding authors.}, Kun Kuang\textsuperscript{1}\footnotemark[2]}\\[0.3em]
  \normalfont\small
  \textsuperscript{1}Zhejiang University, \quad
  \textsuperscript{2}University of the Chinese Academy of Sciences,\\
  \textsuperscript{3}Kuaishou Technology, \quad
  \textsuperscript{4}Zhejiang University of Science and Technology\\[0.2em]
  \texttt{\{zhongwenlin, wuyiquan, kuangkun\}@zju.edu.cn}, \texttt{yuankun03@kuaishou.com}
}

\makeatletter
\renewcommand{\headrulewidth}{0pt}
\def\@maketitle{%
  \lhead{}%
  \begin{center}

    \vspace{0.15in}
    \rule{\textwidth}{3pt}\par
    \vspace{0.2in}
    {\LARGE\bfseries\@title\par}
    \vspace{0.12in}
    \rule{\textwidth}{1pt}\par
    \vspace{0.15in}
    {\normalsize\@author\par}
  \end{center}
  \vskip 0.15in minus 0.05in
}
\makeatother

\usepackage{booktabs}
\usepackage{colortbl}

\definecolor{bestbg}{HTML}{FCECEF}
\definecolor{groupbg}{HTML}{F1F1F1}
\definecolor{avgbg}{HTML}{F4F4F4}
\definecolor{gaincolor}{HTML}{009E9E}

\newcommand{\best}[1]{\cellcolor{bestbg}\textbf{#1}}
\newcommand{\second}[1]{\underline{#1}}
\newcommand{\gain}[1]{\,{\color{gaincolor}\scriptsize$\uparrow_{#1}$}}

\begin{document}

\makeatletter
\let\saved@fnsymbol\@fnsymbol
\def\@fnsymbol#1{\ifnum#1=1 \ensuremath{\ddagger}\else\saved@fnsymbol{#1}\fi}
\makeatother
\maketitle
\makeatletter
\let\@fnsymbol\saved@fnsymbol
\makeatother

\ifdefined\cleanpreview
  \iclrfinaltrue
  \fancyhead{}
\fi


\begin{abstract}
Agentic reinforcement learning (RL) enables large language models (LLMs) to learn from multi-turn interactions with verifiable outcomes. Textual skills can improve exploration by distilling prior interactions into reusable guidance. However, existing skill-based RL approaches often either assume skill injection is beneficial or rely on indirect feedback or rollout-intensive evaluation to assess its utility. Our pilot studies show that rollout-based utility estimation for skill credit assignment is costly and unstable under limited sampling, while skill benefits vary across candidates and depend on the current policy's competence on each query. To address these challenges, we introduce \method, a policy-adaptive training scaffold that combines low-cost skill-utility estimation with selective skill injection. By reusing outcome-verified training trajectories, \method generates candidate skills and estimates their utility through outcome-aware likelihood comparisons, without candidate-specific environment rollouts. The resulting utility guides whether and which skill to inject, retaining skill-free execution when no candidate is estimated to help. Joint training enables the Generator to produce skills tailored to the evolving Executor. Scaffolded RL encourages skill internalization for skill-free inference. Experiments on ALFWorld, WebShop, ScienceWorld, and $\tau^2$-Bench demonstrate gains over strong agentic-RL and skill-based RL baselines across both short- and long-horizon tasks. Ablations and further analyses support the importance of utility-based selection and abstention.
Our code is available at \url{https://github.com/Vaidurya00/SkillScaffold}.


\end{abstract}

\section{Introduction}
\label{sec:introduction}

Agentic reinforcement learning (RL) enables language-model agents to improve
through multi-turn interaction with verifiable environments, but long horizons
and sparse outcome rewards make successful behavior difficult to discover and
reinforce \citep{deepseekmath2024grpo,dong2025arpo}.
Textual skills can ease this challenge by compressing trajectories or feedback
into reusable procedural guidance for subsequent decisions
\citep{shinn2023reflexion,zhao2023expel}.
Existing approaches extract, refine, and retrieve such guidance using external
memory or teacher models
\citep{nottingham2024sso,fu2024autoguide,chen2024automanual,wang2025awm,
zhou2026inspo,fang2025memp}.
SkillRL, for example, uses a stronger teacher to build an evolving SkillBank,
turning sparse trajectory-level feedback into guidance for subsequent
exploration \citep{skillrl2026}.

\begin{figure}[t]
    \vspace{-0.65em}
    \centering
    \includegraphics[width=\linewidth]{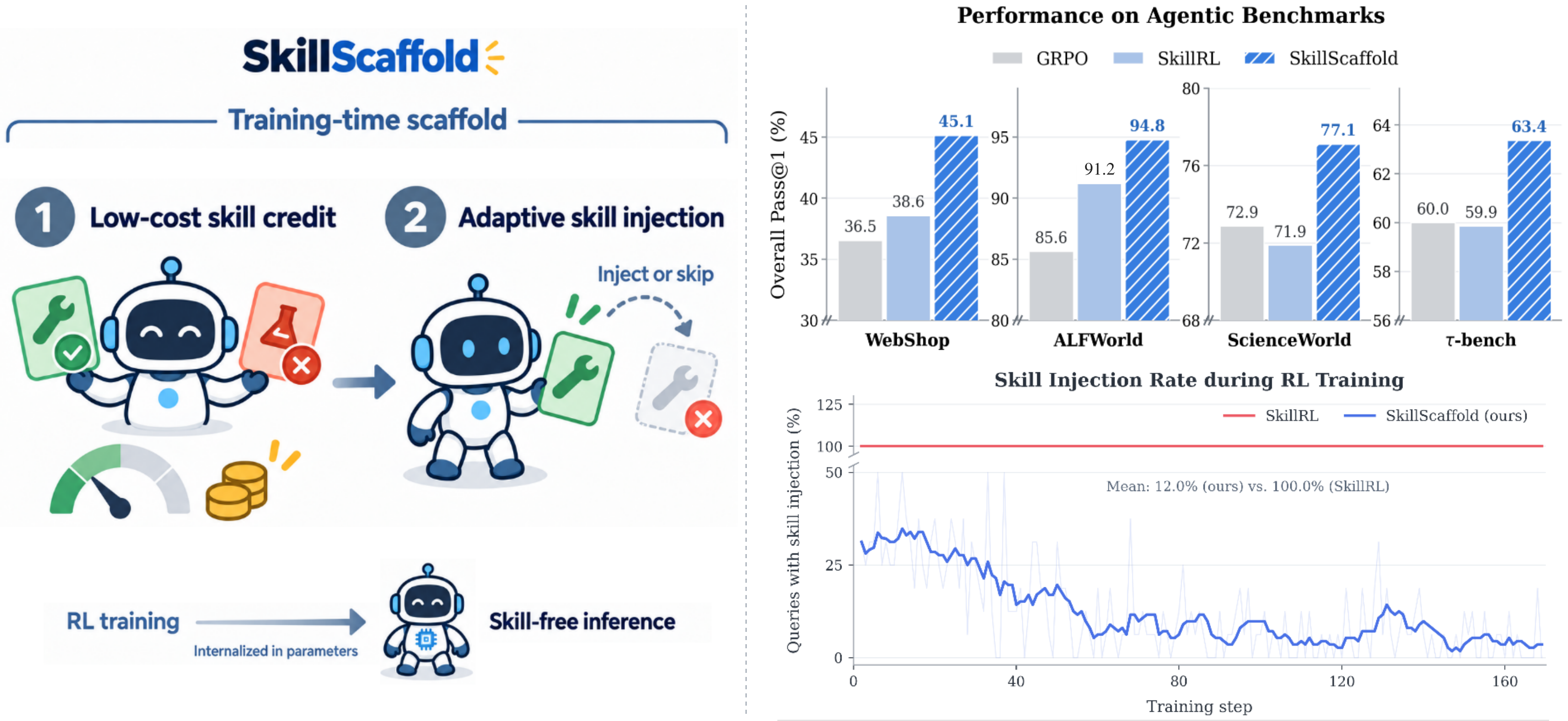}
    \caption{\textbf{Overview and empirical summary of \method.}
    Low-cost, policy-relative skill-utility estimates guide selective
    skill injection during training, while inference remains skill-free.
    Across four agentic benchmarks, \method consistently
    outperforms GRPO and SkillRL while injecting skills for only a small
    fraction of training queries.}
    \label{fig:intro-overview}
    \vspace{-0.8em}
\end{figure}

Recent work shows that a shared policy can generate skills from experience
and use them for subsequent problem solving, allowing generation and execution
capabilities to co-evolve
\citep{wang2025sage,skillrl2026,shi2026skill1,fan2026evolvingrl,
zhou2026inspo,zhang2026sapo,he2026reskill}.
Complementary work internalizes skill-conditioned behavior via curriculum
learning or self-distillation, enabling skill-free inference
\citep{lu2026skill0,wang2026skillsd}.
Building on these directions, we study skills as
\emph{policy-adaptive training scaffolds}, focusing on evaluating and
selectively using guidance for the current policy and query.

Under this training-time scaffold view, \emph{skill credit assignment} uses a
scalar \emph{skill utility} for each candidate to determine whether any skill
should be used and which one to select.
Existing approaches either use generated skills without testing their
benefit \citep{skillrl2026,zhang2026sapo}, rely on indirect and delayed feedback
\citep{shi2026skill1,zhang2026sapo}, or estimate utility through additional
skill-conditioned environment rollouts
\citep{fan2026evolvingrl,he2026reskill,shen2026slim}.
Our pilot studies show that rollout-based estimates are costly and unstable
under limited sampling, while skill benefits vary across candidates and with
how reliably the current policy solves each query without skills
\citep{dong2026harmfulskills}.
These findings motivate a low-cost, policy-relative utility signal for
candidate ranking and abstention.

To address these challenges, we propose \method, a policy-adaptive training
scaffold for agentic RL, summarized in Figure~\ref{fig:intro-overview}.
Its core idea is to replace candidate-specific environment evaluation with
outcome-aware likelihood comparisons on existing training trajectories.
A shared policy serves as both an \emph{Executor}, which collects skill-free
Scout trajectories, and a \emph{Generator}, which produces candidate skills
from outcome-labeled Scout evidence.
For each candidate, \method rescores fixed Scout evidence under candidate-skill
and no-skill contexts to obtain a policy-relative utility estimate.
This proxy favors increased likelihood of successful behavior and reduced
likelihood of observed failures, according to the available outcome evidence.
The highest-utility candidate guides fresh Executor rollouts only when its
estimated utility is positive; otherwise, execution remains skill-free.
Environment rewards supervise the Executor, while skill utility supervises
the Generator through role-specific GRPO objectives, enabling the Generator
to adapt its guidance to the evolving Executor.

Skill-free Scouts and selectively scaffolded Executor trajectories jointly
update the shared policy, encouraging skill internalization.
At deployment, the Generator and the entire skill pathway are removed.
We evaluate \method on ALFWorld, WebShop, ScienceWorld, and $\tau^2$-Bench,
covering diverse multi-turn environments.
Across these settings, \method improves over strong agentic-RL and
skill-based RL baselines under fully skill-free inference.
Ablations and further analyses support the importance of utility-based
selection and abstention.

In summary, the key contributions of this work are as follows:
\begin{itemize}[leftmargin=1.2em, labelsep=0.5em]
    \item We identify two key challenges in skill credit assignment:
    rollout-based evaluation is costly and unstable, while skill benefits
    depend on both the candidate and how reliably the current policy can solve
    the query without skills, motivating candidate ranking and abstention.

    \item We propose \method, a shared-policy training scaffold that
    estimates candidate utility through outcome-aware likelihood
    comparisons on existing trajectories. The resulting signal guides
    skill selection, abstention, and Generator learning without
    candidate-specific environment rollouts.

    \item Experiments on ALFWorld, WebShop, ScienceWorld, and
    $\tau^2$-Bench demonstrate improvements over strong agentic-RL
    and skill-based RL baselines under skill-free inference.
    Ablations and further analyses support the importance of
    utility-based selection and abstention.
\end{itemize}

\section{Preliminaries}
\label{sec:preliminary}
\label{sec:formulation}

\paragraph{Agentic reinforcement learning with skills.}
We consider an LLM policy $\pi_\theta$ that interacts with an environment
over multiple turns through a tool engine $\mathcal{T}$.
Given a task $x\sim\mathcal{D}$ and an optional textual skill $s$,
the Executor generates an interaction trajectory
$\tau=(a_0,o_0,\ldots,a_K,o_K)$ conditioned on $(x,s)$, where
\begin{equation}
  a_t \sim
  \pi_\theta^{\mathrm{exec}}
  (\cdot \mid x,s,\tau_{<t};\mathcal{T}),
  \qquad
  o_t=\mathcal{T}(a_t).
  \label{eq:skill-conditioned-execution}
\end{equation}
Here $s=\varnothing$ denotes skill-free execution.
A rule-based Outcome Verifier assigns an outcome reward $R(\tau;x)$,
and the Executor is optimized to maximize expected environment return:
\begin{equation}
  \max_\theta J_{\mathrm{env}}(\theta)
  =
  \mathbb{E}_{
  x\sim\mathcal{D},\,
  \tau\sim
  \pi_\theta^{\mathrm{exec}}(\cdot\mid x,s;\mathcal{T})
  }
  [R(\tau;x)].
  \label{eq:environment-objective}
\end{equation}

\paragraph{Group Relative Policy Optimization (GRPO).}
GRPO~\citep{deepseekmath2024grpo} optimizes the policy using
group-relative advantages.
For each task $x$, a group of $G$ trajectories
$\{\tau_i\}_{i=1}^{G}$ is sampled from the old policy, and the policy is
updated with
\begin{equation}
  J_{\mathrm{GRPO}}(\theta)
  =
  \mathbb{E}\!\left[
  \frac{1}{G}
  \sum_{i=1}^{G}
  \frac{1}{|\tau_i|}
  \sum_{t=1}^{|\tau_i|}
  \left[
    \min\!\left(
      \rho_{i,t}\widehat{A}_i,
      \operatorname{clip}
      (\rho_{i,t},1-\epsilon,1+\epsilon)\widehat{A}_i
    \right)
    -
    \beta D_{\mathrm{KL}}^{i,t}
  \right]
  \right].
  \label{eq:grpo-objective}
\end{equation}
Here $\rho_{i,t}$ denotes the token-level importance ratio,
$\widehat{A}_i$ is the group-normalized trajectory advantage,
$\epsilon$ is the clipping threshold, and $\beta$ controls the KL penalty.

\section{Motivating Observations}
\label{sec:motivating-observations}

Skill-based agentic RL must decide whether and which generated skill should
guide subsequent exploration. Rollout-based utility estimation is costly and
unstable under practical sampling budgets, while skill utility varies across
candidates and with the current policy.
We first characterize these two challenges before introducing \method.

\subsection{Rollout-Based Skill-Utility Estimation Is Unstable and Costly}
\label{sec:rollout-credit-observation}

For a task $x$ with $M$ candidate skills
$\mathcal{C}_x=\{s_i\}_{i=1}^{M}$, direct skill utility is the downstream
improvement over skill-free execution:
\begin{equation}
    U_{\mathrm{DR}}(s;x)
    =
    \mathbb{E}_{
        \tau\sim\pi_\theta^{\mathrm{exec}}
        (\cdot\mid x,s;\mathcal{T})
    }
    [R(\tau;x)]
    -
    \mathbb{E}_{
        \tau\sim\pi_\theta^{\mathrm{exec}}
        (\cdot\mid x,\varnothing;\mathcal{T})
    }
    [R(\tau;x)].
    \label{eq:downstream-rollout-utility}
\end{equation}

Estimating each condition with $K$ episodes and sharing the skill-free
baseline requires $(M+1)K$ additional environment rollouts per query.

\begin{wrapfigure}[11]{r}{0.40\columnwidth}
    \vspace{-0.6em}
    \centering
    \includegraphics[width=\linewidth]{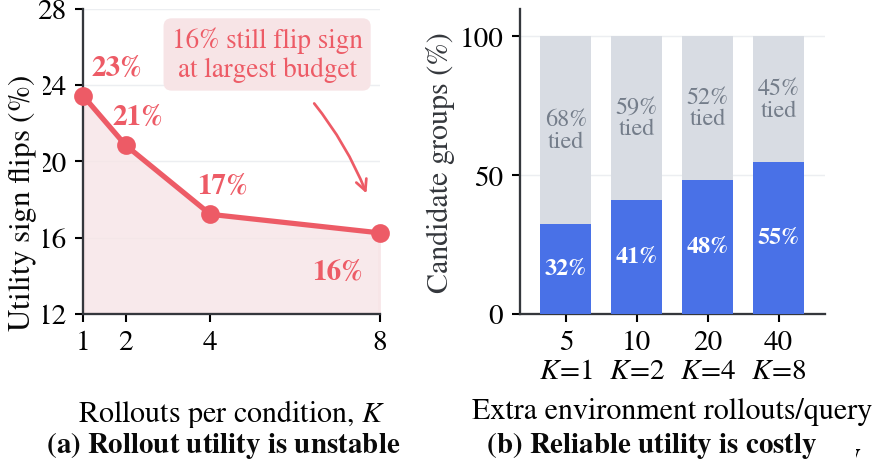}
    \caption{
    \textbf{Rollout-based utility estimation is costly and unstable.}
    (a) Sign disagreement between independent estimates.
    (b) More rollout samples distinguish more candidate groups.
    }
    \label{fig:rollout-credit-pilot}
\end{wrapfigure}

\paragraph{Pilot observation.}
For each candidate--query pair and budget $K$, we form two independent
estimates of $U_{\mathrm{DR}}$ and record a sign disagreement when both are
non-zero but have opposite signs. Figure~\ref{fig:rollout-credit-pilot}(a)
shows that disagreement decreases only from $23\%$ at $K=1$ to $16\%$ at
$K=8$ under sparse outcome rewards.

A query-level candidate group is non-tied when its $M$ candidates do not all
receive the same empirical utility. Figure~\ref{fig:rollout-credit-pilot}(b)
shows that increasing the additional rollout budget eight-fold, from $5$ to
$40$ per query, raises the fraction of non-tied groups only from $32\%$ to
$55\%$, leaving nearly half unresolved.

Online RL must immediately decide \emph{whether} and \emph{which} skill to
inject, whereas reliable candidate-specific estimates require additional
interaction. We therefore seek a low-cost utility estimate that avoids
candidate-specific downstream rollouts.

\subsection{Skill Utility Is Query- and Policy-Dependent}
\label{sec:selective-skill-observation}

Low interaction cost alone is insufficient: skill utility varies across
queries, candidates, and policy competence, so credit assignment must both
rank candidates and abstain when none is helpful.

\begin{wrapfigure}[23]{r}{0.40\columnwidth}
    \captionsetup{skip=2pt}
    \vspace{-0.6em}
    \centering
    \includegraphics[width=\linewidth]{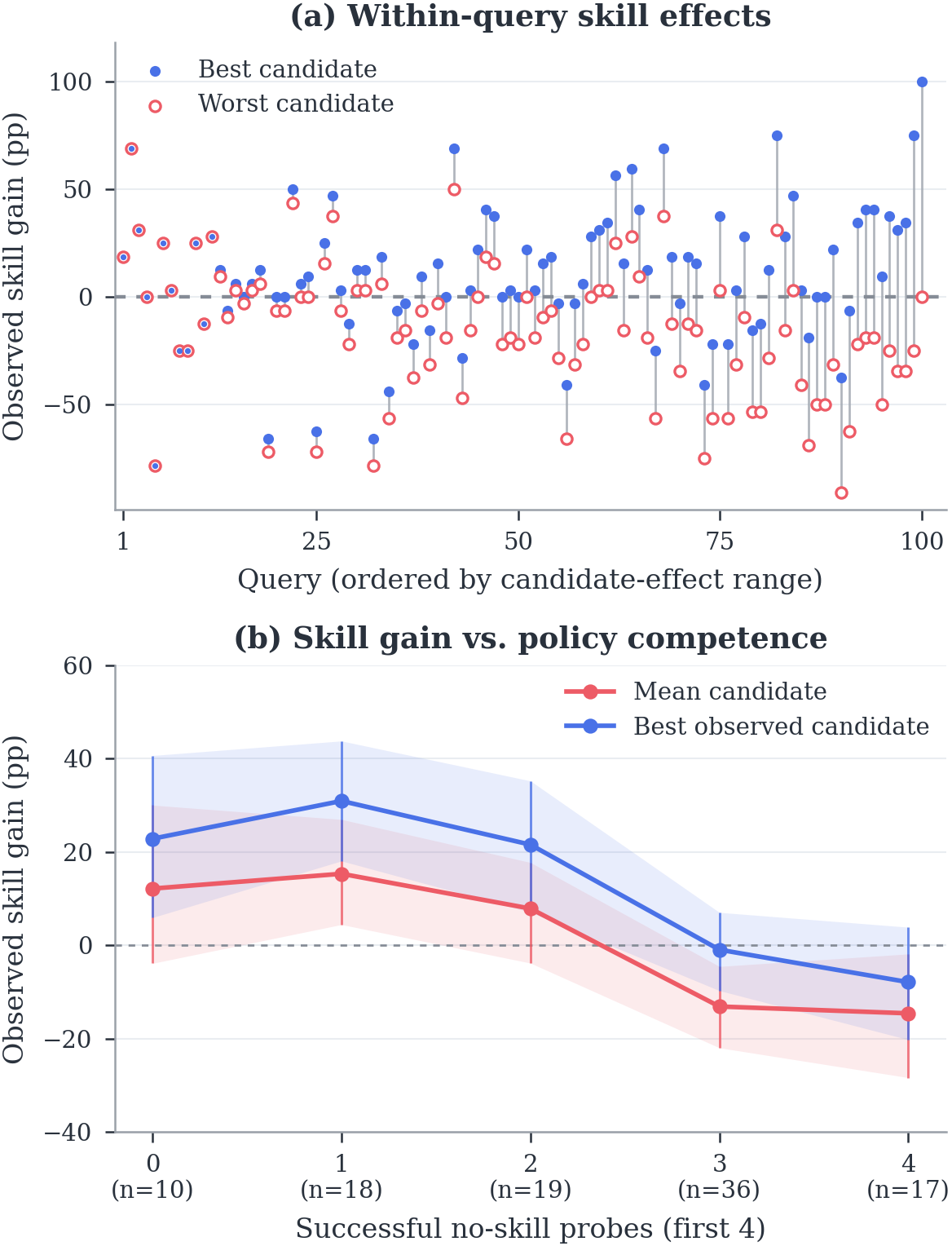}
    \caption{
    \textbf{Skill utility is query- and policy-dependent.}
    (a) Candidate effects vary within a query.
    (b) Skill benefit decreases as skill-free policy competence improves.
    }

    \label{fig:selective-injection-pilot}
\end{wrapfigure}

\paragraph{Within-query variation.}
On a non-saturated pilot subset where candidate differences remain measurable,
Figure~\ref{fig:selective-injection-pilot}(a) compares the best and worst
observed candidates for each query, as measured by downstream rollouts relative
to skill-free execution. Among $100$ queries, $28$ contain both positive- and
negative-effect candidates, showing that generating plausible skills is
insufficient; candidates must be distinguished at the query level.

\paragraph{Dependence on policy competence.}
We measure current competence using four independent skill-free probe rollouts
from the same frozen policy. Here, a probe is an evaluation-only rollout used
to assess the policy's current skill-free competence. This avoids conflation
with historical training-time Scout outcomes. We group queries by their number
of successful probes ($0$--$4$). Figure~\ref{fig:selective-injection-pilot}(b)
reports both the mean candidate effect and the best observed candidate. Average
skill gains are positive when only a minority of probes succeed; when most or
all succeed, the mean effect becomes negative and even the best observed
candidate provides little or no additional benefit.

Thus, skill credit assignment requires a \emph{query- and policy-relative
utility} to identify the most helpful candidate and abstain when none provides
positive additional value. These downstream measurements are used only for
diagnosis, not as training signals for \method.

\section{SkillScaffold}
\label{sec:method}

\begin{figure*}[t]
    \centering
    \includegraphics[width=\textwidth]{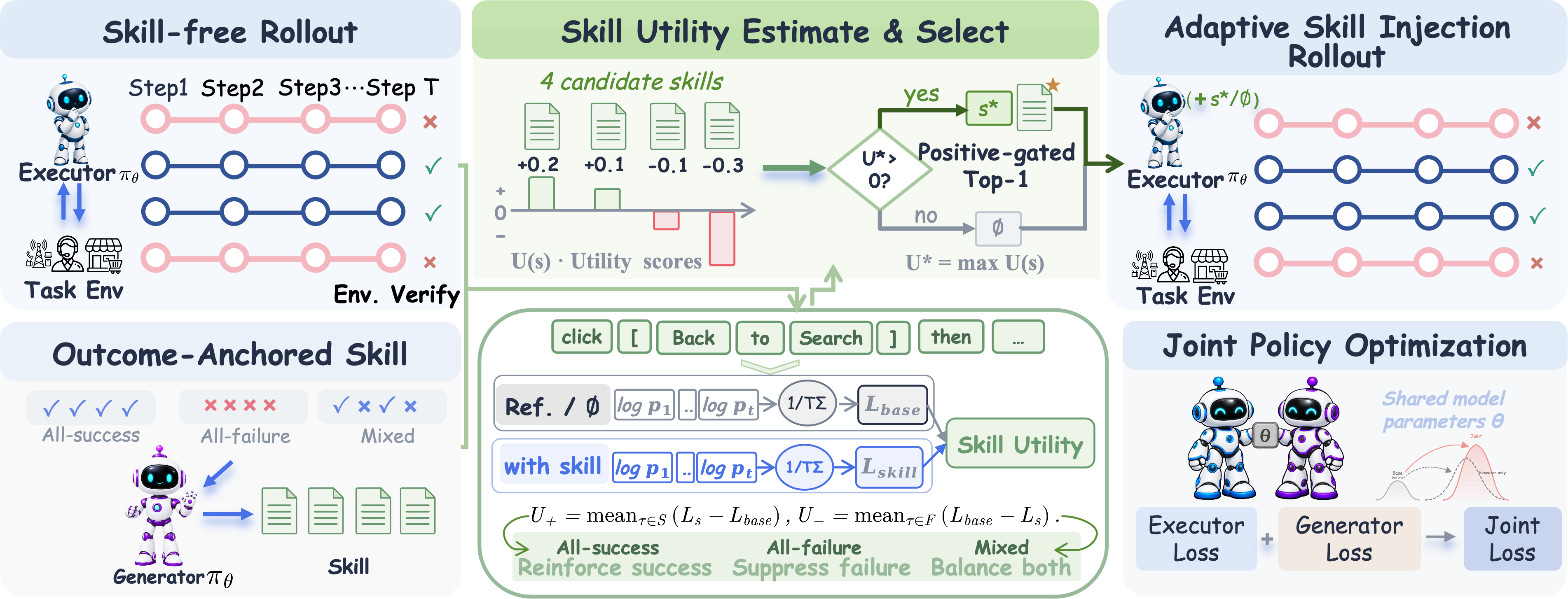}
    \caption{\textbf{Overview of \method.} The five stages show skill-free
    scouting, outcome-anchored generation, utility-based selection, adaptive
    injection, and joint policy optimization.}
    \label{fig:method-overview}
\end{figure*}

\method uses one shared policy $\pi_\theta$ in two roles: the \emph{Executor}
interacts with the environment, while the \emph{Generator} produces candidate
skills from outcome-labeled trajectories. As shown in
Figure~\ref{fig:method-overview}, each update collects skill-free Scouts,
generates and scores candidate skills, and injects only the highest-utility
positive candidate into fresh Executor rollouts. Environment rewards train the
Executor, skill utility trains the Generator, and their shared parameters enable
fully skill-free deployment.

\subsection{Skill-Free Rollout}
\label{sec:skill-free-rollout}

\paragraph{Skill-free Scout collection.}
For task $x$, the current Executor samples $N$ skill-free trajectories:
\begin{equation}
  \tau_i^0 \sim \pi_{\theta}^{\mathrm{exec}}
  (\cdot\mid x,\varnothing;\mathcal{T}),\qquad i=1,\ldots,N.
  \label{eq:scout-rollouts}
\end{equation}
We call these trajectories \emph{Scouts}. An environment verifier assigns
each Scout an outcome $y_i\in\{0,1\}$, which defines
$\mathcal{S}_x=\{i:y_i=1\}$ and $\mathcal{F}_x=\{i:y_i=0\}$.
The resulting group is all-success, all-failure, or mixed.

\subsection{Outcome-Anchored Skill}
\label{sec:four-scout-evidence}

\paragraph{Group-based evidence.}
The Generator receives the task and all outcome-labeled Scouts. It extracts
reusable decisions from all-success groups, diagnoses recurring errors from
all-failure groups, and contrasts successful and failed behavior in mixed
groups. Thus, both successes and failures provide query-specific evidence for
skill generation.

\paragraph{Anchor-based evidence.}
We also consider a lightweight variant that uses one verified successful
Scout as its evidence anchor. When $\mathcal{S}_x\neq\varnothing$, we choose
$a_x=\min\mathcal{S}_x$ and provide $\tau_{a_x}^0$ to the Generator. If no
Scout succeeds, this variant skips generation and keeps the subsequent
Executor rollouts skill-free.

\paragraph{Candidate generation.}
Let $E_x^v$ denote the evidence record for variant
$v\in\{\mathrm{grp},\mathrm{anc}\}$. The Generator independently samples
$M$ candidate skills,
\begin{equation}
  s_j\sim\pi_{\theta}^{\mathrm{gen}}(\cdot\mid E_x^v),
  \qquad \mathcal{C}_x=\{s_j\}_{j=1}^{M}.
  \label{eq:skill-generation}
\end{equation}
Each candidate is a structured skill card containing applicability
conditions, an executable procedure, conditional decision rules, and
verification, recovery, or stopping criteria. Format-invalid outputs are
excluded from selection.

\subsection{Skill Utility Estimate \& Select}
\label{sec:utility}

A skill credit-assignment mechanism must assign each candidate a scalar
\emph{skill utility} $U(s;x)$. A useful skill should be compatible with verified
successful behavior and incompatible with observed failed behavior. Directly testing this criterion
with fresh environment rollouts would reproduce the cost and variance
identified in Section~\ref{sec:rollout-credit-observation}. Instead, we keep
the Scouts fixed and compare how the current policy scores the same behavior
with and without each candidate skill.

\paragraph{Paired trajectory scoring.}
Write the Executor-generated reasoning and action tokens in a fixed Scout
trajectory $\tau$ as $u_{1:T}$. For skill context $c$, define
\begin{equation}
  \ell_{\theta}(c;\tau)
  =
  \frac{1}{T}
  \sum_{t=1}^{T}
  \log\pi_{\theta}^{\mathrm{exec}}
  (u_t\mid x,c,\tau_{<t}),
  \qquad
  d(s,\tau)=
  \ell_{\theta}(s;\tau)-\ell_{\theta}(\varnothing;\tau).
  \label{eq:trajectory-likelihood}
\end{equation}
Here $\tau_{<t}$ is the recorded interaction prefix preceding $u_t$, including
previous Executor outputs and environment observations. A positive
$d(s,\tau)$ means that conditioning on $s$ increases the likelihood of the
observed Scout behavior relative to skill-free execution.

\paragraph{Group-based utility.}
We aggregate the trajectory-level differences according to the observed
outcomes:
\begin{equation}
  U^{\mathrm{grp}}(s;x)=
  \begin{cases}
    \displaystyle
    \frac{1}{|\mathcal{S}_x|}
    \sum_{i\in\mathcal{S}_x}d(s,\tau_i^0),
      & \mathcal{F}_x=\varnothing,\\[6pt]
    \displaystyle
    -\frac{1}{|\mathcal{F}_x|}
    \sum_{i\in\mathcal{F}_x}d(s,\tau_i^0),
      & \mathcal{S}_x=\varnothing,\\[6pt]
    \displaystyle
    \frac{1}{2|\mathcal{S}_x|}
    \sum_{i\in\mathcal{S}_x}d(s,\tau_i^0)
    -
    \frac{1}{2|\mathcal{F}_x|}
    \sum_{i\in\mathcal{F}_x}d(s,\tau_i^0),
      & \text{otherwise}.
  \end{cases}
  \label{eq:scaffold-utility}
\end{equation}
For an all-success group, utility increases when the skill raises the
likelihood of verified successful behavior. For an all-failure group, the
sign is reversed, so utility increases when the skill suppresses observed
failed behavior. For a mixed group, the success-support and
failure-suppression terms receive equal total weight regardless of their
relative counts. We refer to this balanced quantity as the
\emph{group-aware skill utility}.

\paragraph{Anchor-based utility.}
The selected successful Scout supplies the scoring anchor as well as the
generation evidence:
\begin{equation}
  U^{\mathrm{anc}}(s;x)=d(s,\tau_{a_x}^0),
  \qquad \mathcal{S}_x\neq\varnothing.
  \label{eq:anchor-utility}
\end{equation}
Anchor-based utility uses one successful Scout, whereas group-based utility
also incorporates failure evidence.

Skill utility provides a policy-relative signal for candidate ranking and
abstention. By comparing outcome-conditioned likelihoods over verified Scout
trajectories, it identifies skills aligned with the current Executor, while
environment outcomes supervise Executor learning. Appendix~\ref{app:theory}
provides a log-likelihood-ratio interpretation and corresponding selection
guarantees.

\paragraph{Positive-gated selection.}
Let $U_j=U^v(s_j;x)$, and let $\mathcal{V}_x$ contain candidates
that satisfy the required format and have a finite utility. If
$\mathcal{V}_x$ is nonempty, we define
$j^*=\arg\max_{j\in\mathcal{V}_x}U_j$ and select
\begin{equation}
  s_x^*=
  \begin{cases}
    s_{j^*}, & \mathcal{V}_x\neq\varnothing\ \text{and}\ U_{j^*}>0,\\
    \varnothing, & \text{otherwise}.
  \end{cases}
  \label{eq:positive-gated-selection}
\end{equation}
The ranking identifies the most compatible candidate, while the sign test
determines whether any skill should be injected.

\subsection{Adaptive Skill Injection Rollout}
\label{sec:adaptive-injection}

Given the selected context $s_x^*$, the Executor samples $L$ fresh
\emph{Post-Scout} trajectories,
$\tau_k^1\sim\pi_{\theta}^{\mathrm{exec}}
(\cdot\mid x,s_x^*;\mathcal{T})$, $k=1,\ldots,L$. Thus, every query proceeds
through the same Executor interface: a positive-utility skill is supplied when
available, and $s_x^*=\varnothing$ otherwise.

\subsection{Joint Policy Optimization}
\label{sec:joint-optimization}

\paragraph{Role-specific learning signals.}
The $N$ Scouts and $L$ post-selection trajectories form one Executor GRPO
group whose rewards come exclusively from the environment verifier. The $M$
generated skill texts form a separate Generator GRPO group.
Their detached rewards are derived from skill utility:
\begin{equation}
  r_j^{\mathrm{gen}}=
  \begin{cases}
    \max\{0,\operatorname{sg}[U_j]\}, & v=\mathrm{anc},\\
    \operatorname{sg}[U_j], & v=\mathrm{grp},
  \end{cases}
  \label{eq:generator-reward}
\end{equation}
Because anchor evidence contains only successes, nonpositive utilities are
clipped to zero; group-based rewards retain their signs to distinguish
success-supporting from failure-reinforcing candidates. Executor and Generator
rewards are normalized separately.

We optimize the shared parameters with
\begin{equation}
  J_{\mathrm{SS}}(\theta)
  =
  J_{\mathrm{GRPO}}^{\mathrm{exec}}(\theta;R)
  +\lambda_{\mathrm{gen}}
   J_{\mathrm{GRPO}}^{\mathrm{gen}}(\theta;r^{\mathrm{gen}})
  +\eta_{\mathrm{gen}}\mathcal{H}_{\mathrm{gen}}(\theta),
  \label{eq:joint-objective}
\end{equation}
where $\lambda_{\mathrm{gen}}$ weights Generator learning and
$\eta_{\mathrm{gen}}$ controls its entropy regularization.
The Executor learns from verified task outcomes, while the Generator learns
from policy-relative utility. Because they share $\theta$, both skill-free
Scouts and selectively scaffolded Post-Scouts update the same policy,
reinforcing useful scaffolded behavior without separate distillation.

\paragraph{Skill-free deployment.}
\label{sec:skill-free-deployment}
At deployment, we disable the skill pathway and use
$\pi_{\theta^*}^{\mathrm{exec}}(\cdot\mid x,\varnothing,h_t;\mathcal{T})$,
eliminating Scout collection and all skill-related inference overhead.


\section{Experiments}
\label{sec:experiments}

\subsection{Experimental Setup}

\paragraph{Benchmarks.}
We evaluate \method on four interactive benchmarks. \textbf{ALFWorld}
\citep{shridhar2021alfworld} is a text-based household environment requiring
multi-step planning and object interaction. \textbf{WebShop}
\citep{yao2022webshop} is an online-shopping simulator in which agents search
for products matching user specifications. \textbf{ScienceWorld}
\citep{wang2022scienceworld} evaluates interactive scientific reasoning and
experimentation. These three are short-horizon settings. In contrast,
\textbf{$\tau^2$-Bench}~\citep{yao2024taubench} evaluates long-horizon
tool-agent-user interaction across Airline, Retail, and Telecom. We report
success rate for the first three benchmarks and Pass@1 for $\tau^2$-Bench.
Dataset and evaluation details are provided in
Appendix~\ref{app:benchmark-details}.

\paragraph{Baselines and Implementation Details.}
All main runs use Qwen3-8B. Baselines include
(1) \textbf{Prompt- and memory-based methods:}
ReAct~\citep{yao2023react}, Reflexion~\citep{shinn2023reflexion},
Mem0~\citep{chhikara2025mem0}, ExpeL~\citep{zhao2023expel},
MemP~\citep{fang2025memp}, and direct skill prompting.
(2) \textbf{RL methods:} GRPO~\citep{deepseekmath2024grpo} and
GiGPO~\citep{feng2025gigpo} optimize the policy without skill augmentation.
(3) \textbf{Skill-based RL methods:} INSPO~\citep{zhou2026inspo} and
SkillRL~\citep{skillrl2026} use teacher-generated guidance, whereas
Skill1~\citep{shi2026skill1}, Evolving-RL~\citep{fan2026evolvingrl}, and
SEED~\citep{seed2026} generate or distill guidance from the policy's own
experience. Full implementation details are provided in
Appendix~\ref{app:implementation}.



\subsection{Main Results}
\label{sec:results}

\begin{table*}[t]
\centering
\caption{\textbf{Performance on WebShop and ALFWorld with Qwen3-8B.} We report
success rates (\%) across task categories and their averages. $\dagger$ denotes
evaluation with skills, while $*$ denotes evaluation without skills.
{\setlength{\fboxsep}{0pt}\colorbox{bestbg}{\textbf{Best}}} and
\underline{second-best} results are marked
accordingly.}
\label{tab:webshop}
\label{tab:alfworld}
\begingroup
\footnotesize
\setlength{\tabcolsep}{0.35pt}
\renewcommand{\arraystretch}{0.96}
\newcolumntype{Y}{>{\centering\arraybackslash}p{2.35em}}

\resizebox{\textwidth}{!}{%
\begin{tabular}{l*{10}{Y}|*{7}{Y}}
\toprule
& \multicolumn{10}{c}{\textbf{WebShop}}
& \multicolumn{7}{c}{\textbf{ALFWorld}} \\
\cmidrule(lr){2-11}\cmidrule(lr){12-18}

& \multicolumn{5}{c}{ID}
& \multicolumn{4}{c}{OOD}
& \multicolumn{1}{c}{\multirow{2}{*}{Avg.}}
& \multicolumn{6}{c}{Task Type}
& \multicolumn{1}{c}{\multirow{2}{*}{Avg.}} \\
\cmidrule(lr){2-6}\cmidrule(lr){7-10}
\cmidrule(lr){12-17}

\multicolumn{1}{c}{Method}
& App. & Ele. & Foo. & Oth. & ID
& Acc. & Bea. & Hom. & OOD &
& Pic. & Hea. & Coo. & Cle. & Loo. & Pk2. & \\
\midrule

ReAct
& 18.8 & 16.9 & 18.1 & 17.0 & 17.6
& 14.1 & 16.0 & 8.6 & 12.6 & \cellcolor{avgbg}16.1
& 59.4 & 32.1 & 41.1 & 29.8 & 41.0 & 5.2 & \cellcolor{avgbg}35.6 \\

\midrule
\rowcolor{groupbg}
\multicolumn{11}{l|}{\textit{Prompt-based Agentic / Memory-based Methods}}
& \multicolumn{7}{c}{} \\

Reflexion
& 22.7 & 19.5 & 20.8 & 21.1 & 20.9
& 16.7 & 24.5 & 13.0 & 17.5 & \cellcolor{avgbg}19.8
& 59.9 & 21.7 & 42.9 & 41.5 & 38.9 & 25.7 & \cellcolor{avgbg}39.3 \\

Mem0
& 18.3 & 15.7 & 14.5 & 17.0 & 16.4
& 13.3 & 15.3 & 8.6 & 12.1 & \cellcolor{avgbg}15.1
& 37.0 & 25.0 & 41.7 & 41.9 & 30.6 & 29.4 & \cellcolor{avgbg}35.0 \\

ExpeL
& 25.2 & 22.5 & 26.5 & 24.3 & 24.4
& 17.3 & 20.4 & 9.6 & 15.4 & \cellcolor{avgbg}21.6
& 32.3 & 48.9 & 47.6 & 37.5 & 41.7 & 29.4 & \cellcolor{avgbg}39.7 \\

MemP
& 23.8 & 18.3 & 22.1 & 18.4 & 20.4
& 13.6 & 13.9 & 9.9 & 12.4 & \cellcolor{avgbg}17.9
& 53.1 & 46.2 & 44.1 & 55.2 & 29.2 & 32.4 & \cellcolor{avgbg}45.2 \\

w/ Skill
& 12.5 & 13.2 & 17.7 & 14.9 & 14.3
& 11.1 & 18.5 & 6.3 & 11.4 & \cellcolor{avgbg}13.4
& 56.3 & 8.2 & 43.5 & 45.6 & 43.1 & 16.9 & \cellcolor{avgbg}36.8 \\

\midrule
\rowcolor{groupbg}
\multicolumn{11}{l|}{\textit{RL Methods}}
& \multicolumn{7}{c}{} \\

GRPO
& 36.2 & 36.4 & 36.8 & 44.6 & 38.2
& 29.2 & 43.8 & 28.8 & 32.9 & \cellcolor{avgbg}36.5
& 97.4 & 86.4 & 92.3 & 82.3 & 71.5 & 80.9 & \cellcolor{avgbg}85.6 \\

GiGPO
& 33.7 & \second{46.6} & 28.5 & 52.8 & 41.2
& \best{39.4} & \second{54.4} & 30.0 & \second{40.0} & \cellcolor{avgbg}40.8
& 87.0 & 82.1 & 92.3 & 75.8 & 61.1 & 53.7 & \cellcolor{avgbg}76.7 \\

\midrule
\rowcolor{groupbg}
\multicolumn{11}{l|}{\textit{Skill-based RL Methods}}
& \multicolumn{7}{c}{} \\

INSPO$^{\dagger}$
& 33.6 & 40.9 & 34.7 & 47.4 & 39.3
& 30.8 & 46.3 & 30.8 & 34.8 & \cellcolor{avgbg}37.9
& 82.3 & 73.4 & 91.1 & 70.2 & 62.5 & 58.8 & \cellcolor{avgbg}73.7 \\

SkillRL$^{\dagger}$
& 31.9 & 37.3 & \second{38.6} & 50.3 & 39.1
& 36.7 & 46.5 & 31.5 & 37.4 & \cellcolor{avgbg}38.6
& 84.7 & 82.8 & 90.2 & \second{95.8} & \best{98.6}
& \second{96.8} & \cellcolor{avgbg}91.2 \\

SKILL1$^{\dagger}$
& 33.5 & 36.6 & 28.0 & 48.3 & 36.7
& 29.7 & 53.9 & 28.4 & 35.6 & \cellcolor{avgbg}36.4
& 91.7 & \best{95.7} & 90.5 & 83.9 & \second{94.4}
& \best{100.0} & \cellcolor{avgbg}\second{91.8} \\

Evolving-RL$^{\dagger}$
& \best{41.7} & 41.1 & 36.4 & 52.0 & \second{42.7}
& 36.1 & \best{55.8} & 31.7 & 39.7 & \cellcolor{avgbg}\second{41.8}
& 92.2 & 71.2 & 76.2 & 61.3 & 66.0 & 87.5 & \cellcolor{avgbg}74.8 \\

SEED$^{*}$
& 34.4 & 41.9 & 30.6 & \second{53.4} & 40.3
& 36.6 & 52.1 & \second{33.2} & 39.4 & \cellcolor{avgbg}40.1
& 84.4 & 83.7 & 89.3 & 92.7 & 83.3 & 77.2 & \cellcolor{avgbg}85.9 \\

Evolving-RL$^{*}$
& 26.2 & 31.7 & 25.0 & 29.8 & 28.6
& 21.7 & 39.4 & 20.2 & 25.8 & \cellcolor{avgbg}27.7
& \second{99.0} & 70.1 & 79.2 & 57.7 & 85.4 & 78.7 & \cellcolor{avgbg}77.0 \\

\addlinespace[2pt]
\ours$^{*}$
& \second{38.3} & \best{47.6} & \best{42.7} & \best{57.9}
& \best{46.6} & \second{37.7} & \best{55.8} & \best{36.5}
& \best{42.0} & \best{45.1}
& \best{100.0} & \second{87.0} & \best{100.0} & \best{96.8}
& 88.9 & 94.1 & \best{94.8} \\

\bottomrule
\end{tabular}%
}
\endgroup
\end{table*}

\paragraph{Overall performance.}
Tables~\ref{tab:webshop} and~\ref{tab:secondary} show a clear progression over
increasingly competitive baselines. First, \method substantially outperforms
prompt- and memory-based methods, with gains of 23.5 points on WebShop and
49.6 on ALFWorld over their strongest results. It then surpasses standard RL
methods across all four benchmarks, including gains of 9.2 points on ALFWorld
and 4.2 on ScienceWorld. Finally, \method remains best against skill-based RL
methods on every benchmark, improving the strongest such results by 3.3 points
on WebShop and 3.0 on ALFWorld. Its 6.3-point gain on Telecom further extends
this advantage to long-horizon tool use. Full training dynamics are provided
in Appendix~\ref{app:additional-results}.

\paragraph{Skill-free deployment and generalization.}
\method removes the Generator and all textual skills at inference, yet
outperforms skill-conditioned baselines on both WebShop and ALFWorld. On
WebShop, it reaches 45.1 without skills, compared with 41.8 for Evolving-RL
with skills and 27.7 after its skills are removed. The gains also hold across
ID and OOD splits: \method reaches 46.6/42.0 on WebShop and 96.2/48.8 on
ScienceWorld, exceeding the strongest competing results on both splits.
Thus, the improvements do not rely on retaining an external skill pathway and
generalize beyond the training categories. Full Pass@$k$ results are provided
in Appendix Table~\ref{tab:tau2-full}.

\begin{table*}
\centering
\caption{{\bfseries\boldmath Performance on ScienceWorld and $\tau^2$-Bench with
Qwen3-8B.} We report success rates (\%) on the ScienceWorld ID/OOD splits and
the three $\tau^2$-Bench domains. Teal arrows indicate absolute gains in
percentage points over the strongest competing result.
{\setlength{\fboxsep}{0pt}\colorbox{bestbg}{\textbf{Best}}} and
\underline{second-best} results are marked
accordingly.}
\label{tab:scienceworld}
\label{tab:secondary}
\begingroup
\footnotesize
\setlength{\tabcolsep}{3.0pt}

\scalebox{0.90852}{%
\begin{tabular}{lccc|cccc}
\toprule
& \multicolumn{3}{c|}{\textbf{ScienceWorld}}
& \multicolumn{4}{c}{\textbf{$\tau^2$-Bench}} \\
\cmidrule(lr){2-4}\cmidrule(lr){5-8}

\multicolumn{1}{c}{Method}
& ID & OOD & Avg.
& Airline & Retail & Telecom & Avg. \\
\midrule

ReAct
& 19.3 & 11.1 & \cellcolor{avgbg}16.0
& 17.8 & 35.5 & 7.6 & \cellcolor{avgbg}20.3 \\

SFT
& -- & -- & \cellcolor{avgbg}--
& 41.0 & 40.1 & 56.6 & \cellcolor{avgbg}45.9 \\

GRPO
& \second{92.6} & 43.6 & \cellcolor{avgbg}\second{72.9}
& \second{45.8} & 57.2 & \second{77.0} & \cellcolor{avgbg}\second{60.0} \\

SkillRL
& 88.2 & \second{47.7} & \cellcolor{avgbg}71.9
& \best{46.0} & \second{57.9} & 75.7 & \cellcolor{avgbg}59.9 \\

\ours
& \best{96.2\gain{3.6}} & \best{48.8\gain{1.1}} & \best{77.1\gain{4.2}}
& \best{46.0} & \best{60.8\gain{2.9}} & \best{83.3\gain{6.3}}
& \best{63.4\gain{3.4}} \\

\bottomrule
\end{tabular}%
}
\endgroup
\end{table*}

\FloatBarrier


\subsection{Ablation Studies}
\label{sec:ablation}

\begin{wraptable}{r}{0.49\textwidth}
\vspace{-0.8em}
\centering
\caption{\textbf{Ablation results on ALFWorld and WebShop.}}
\label{tab:ablation}
\begingroup
\small
\setlength{\tabcolsep}{3pt}
\renewcommand{\arraystretch}{1.02}

\begin{tabular}{@{}p{0.65\linewidth}
                >{\centering\arraybackslash}p{0.13\linewidth}
                >{\centering\arraybackslash}p{0.13\linewidth}@{}}
\toprule
\multicolumn{1}{c}{Variant} & ALF. & Web. \\
\midrule

\rowcolor{groupbg}
\ours (Full)
& \textbf{94.8} & \textbf{45.1} \\

\midrule
\multicolumn{3}{@{}l}{\textit{Generator Learning}} \\
\quad w/o Generator learning
& 89.6 & 39.5 \\
\quad No Gen. loss + random selection
& 81.3 & 40.3 \\

\midrule
\multicolumn{3}{@{}l}{\textit{Utility Selection}} \\
\quad Random selection
& 79.9 & 39.2 \\
\quad Shuffled utility
& 82.8 & 37.1 \\

\midrule
\multicolumn{3}{@{}l}{\textit{Selective Scaffolding}} \\
\quad w/o abstention
& 85.8 & 39.6 \\
\quad No Executor injection
& \second{90.3} & \second{41.5} \\

\bottomrule
\end{tabular}
\endgroup
\vspace{-0.2em}
\end{wraptable}

Table~\ref{tab:ablation} isolates utility-supervised Generator learning,
utility-based candidate selection, and positive-gated skill injection.
Removing Generator learning reduces ALFWorld/WebShop performance to
89.6/39.5, showing that skill generation must adapt to the evolving policy.
Random selection and shuffled candidate--utility assignments yield 79.9/39.2
and 82.8/37.1, showing that utility must both guide selection and remain paired
with the candidate it evaluates. Removing the positive gate gives 85.8/39.6,
confirming that even the top-ranked skill can be harmful when its utility is
nonpositive. Finally, training the Generator without injecting selected skills
into Executor rollouts gives 90.3/41.5, showing that Generator supervision
alone cannot replace skill-scaffolded exploration.


\subsection{Validating Skill Utility for Credit Assignment}
\label{sec:skill-credit-validation}

\begin{wrapfigure}[12]{r}{0.5\textwidth}
\vspace{-0.7em}
\centering
\includegraphics[width=\linewidth]
{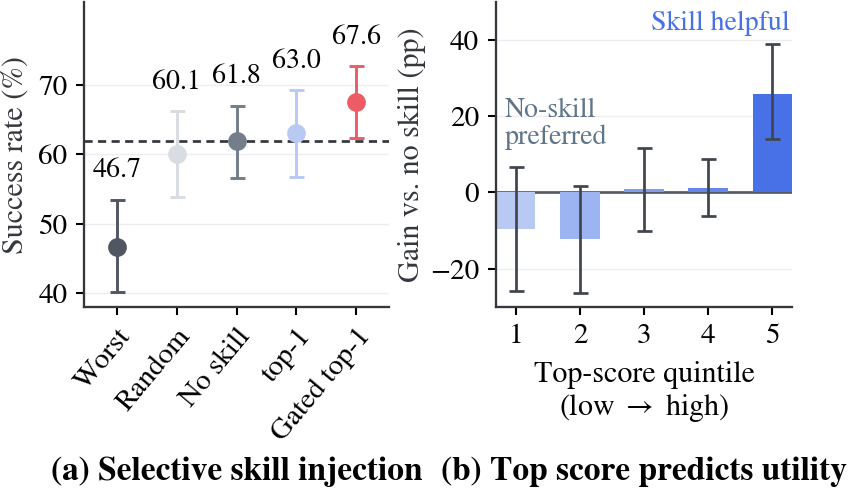}
\captionsetup{skip=0pt}
\caption{\textbf{Skill-utility validation.}}
\label{fig:skill-credit-validation}
\vspace{2em}
\end{wrapfigure}

We reuse the independently evaluated query--candidate sets from the pilot
study. For each candidate, the likelihood-based utility is computed only from
the fixed Scout trajectories, without using its downstream evaluation
rollouts. Figure~\ref{fig:skill-credit-validation}(a) applies four decision
rules to the same candidate sets: skill-free execution, random injection,
forced injection of the utility-ranked top-1 candidate, and top-1 injection
only when its utility is positive. Their success rates are 61.8\%, 60.1\%,
63.0\%, and 67.6\%, respectively. This shows that ranking alone provides only
a modest gain, whereas positive gating is critical for avoiding unhelpful or
harmful injections. Panel (b) partitions candidate--query pairs into five
utility quintiles and uses independent downstream rollouts only for evaluation,
reporting gain over skill-free execution. The measured gain rises with utility,
supporting its use for both ranking and abstention.


\subsection{SkillScaffold Improves Training-Time Exploration}
\label{sec:exploration-analysis}

\begin{figure*}[t]
    \centering
    \includegraphics[width=0.98\textwidth]
    {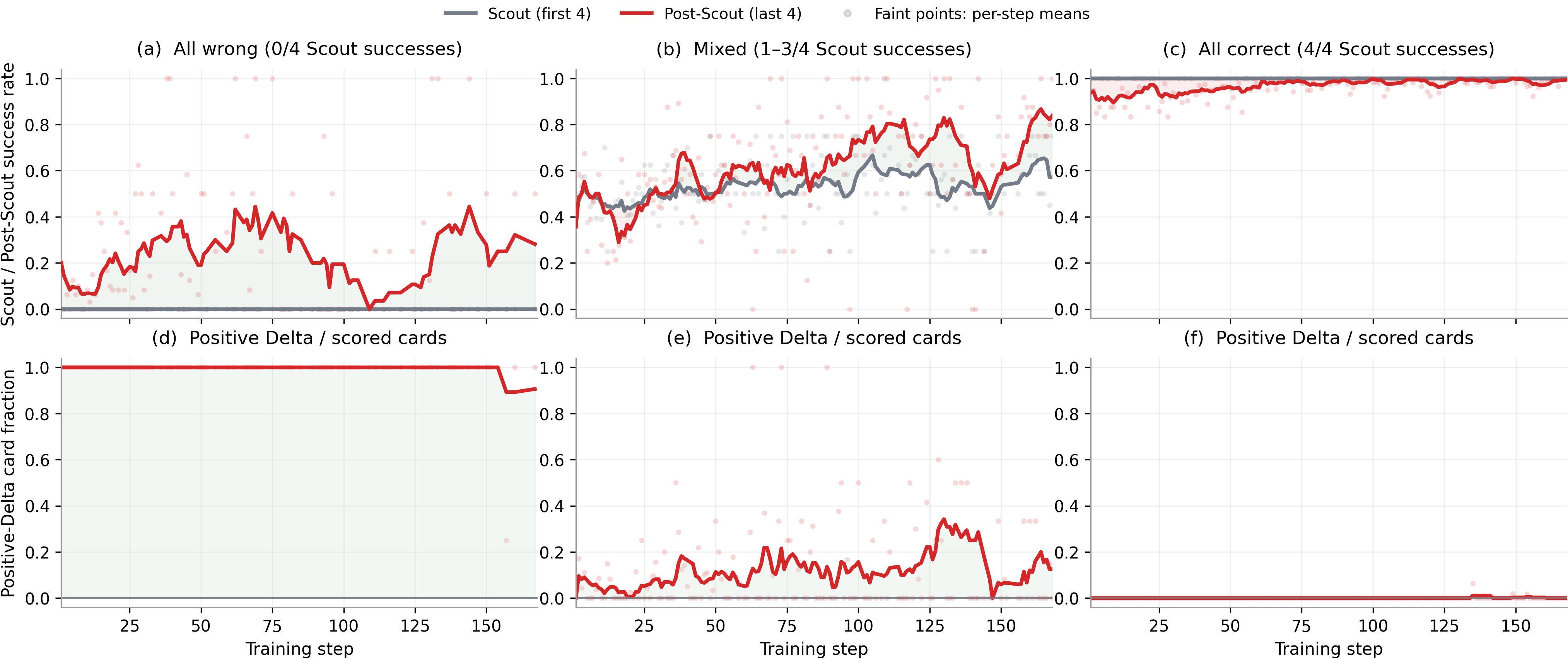}
    \caption{
    \textbf{ALFWorld exploration dynamics.}
    Scout/Post-Scout success (top) and skill-injection rates (bottom), grouped
    by all-wrong, mixed, and all-correct Scout outcomes.
    }
    \label{fig:alfworld-exploration}
\end{figure*}

Figure~\ref{fig:alfworld-exploration} shows that scaffolding concentrates on
queries where skill-free exploration is weakest. Relative to Scouts,
Post-Scout success improves by 21.1 points for all-wrong groups. This indicates
that selected skills can unlock successful exploration even when skill-free
Scouts discover no solution. The gain is 5.4 points for mixed groups, while
performance changes little for all-correct groups ($-2.3$ points).
Correspondingly, the gate injects skills almost universally for all-wrong
groups, selectively for mixed groups, and rarely for all-correct groups.
Cross-benchmark curves are reported in Appendix~\ref{app:additional-results}.


\subsection{Skill Internalization into Model Parameters}
\label{sec:skill-internalization}

\begin{wraptable}{r}{0.49\textwidth}
    \vspace{-0.7em}
    \centering
    \caption{
    \textbf{Skill-free evidence of internalization.}
    Later-minus-early skill-free success on fixed early skill-assisted
    training cohorts, with 95\% bootstrap intervals.
    }
    \label{tab:skill-free-recovery}
    \begingroup
    \footnotesize
    \setlength{\tabcolsep}{3pt}
    \renewcommand{\arraystretch}{1.04}
    \begin{tabular}{@{}lcc@{}}
        \toprule
        Benchmark & $N$ & Recovery (pp) \\
        \midrule
        WebShop
        & 19
        & $+19.7\ [0.0,\ 39.5]$ \\

        ALFWorld
        & 54
        & $+29.6\ [20.4,\ 38.9]$ \\

        ScienceWorld
        & 50
        & $+65.0\ [56.0,\ 73.5]$ \\

        $\tau^2$-Bench
        & 98
        & $+16.6\ [10.7,\ 22.5]$ \\
        \bottomrule
    \end{tabular}
\endgroup
\vspace{-0.8em}
\end{wraptable}

Table~\ref{tab:skill-free-recovery} tests whether strategies first exposed
through selected skills become encoded in the shared model parameters. We fix
cohorts of training queries that benefit from skill-assisted exploration at an
early checkpoint and reevaluate them at later checkpoints with the skill
pathway disabled. Skill-free success improves by 19.7 points on WebShop, 29.6
on ALFWorld, 65.0 on ScienceWorld, and 16.6 on $\tau^2$-Bench; intervals
exclude zero on the latter three benchmarks. Because the later Executor
receives no skill text, these gains
indicate that shared-parameter training transfers skill-guided strategies into
the policy itself. A qualitative internalization case is provided in
Appendix~\ref{app:internalization-case}.


\section{Conclusion}
\label{sec:conclusion}

We introduced \method, a policy-adaptive training scaffold for skill credit
assignment in agentic RL. Rather than assuming generated skills are helpful or
evaluating every candidate through additional environment interaction,
\method reuses outcome-verified trajectories to estimate policy-relative
utility through outcome-aware likelihood comparisons. This signal supports
candidate ranking, abstention, and Generator learning, while environment
outcomes continue to supervise the Executor. Across diverse short- and
long-horizon benchmarks, \method improves over strong agentic-RL and
skill-based RL baselines under fully skill-free inference. Overall, \method
selectively uses skills to improve training-time exploration, absorbs their
guidance into the policy, and removes the skill pathway at deployment.


\subsection*{Reproducibility Statement}

Section~\ref{sec:method} describes \method and its training objective.
Appendices~\ref{app:implementation}--\ref{app:prompts-cases} provide
implementation details, additional results, and complete prompts and examples.
We will provide the implementation code in the supplementary material.

\subsection*{AI Use Statement}

Generative AI tools were used to assist with language editing, LaTeX
formatting, and code navigation. They were not treated as sources of evidence
or used to autonomously determine the experimental design, analysis, or
conclusions. The authors reviewed and verified all AI-assisted text and code
and take full responsibility for the scientific claims and final manuscript.

\bibliography{iclr2027_conference}

@article{yao2023react,
  title={React: Synergizing reasoning and acting in language models},
  author={Yao, Shunyu and Zhao, Jeffrey and Yu, Dian and Du, Nan and Shafran, Izhak and Narasimhan, Karthik and Cao, Yuan},
  journal={arXiv preprint arXiv:2210.03629},
  year={2022}
}

@article{deepseekmath2024grpo,
  title={Deepseekmath: Pushing the limits of mathematical reasoning in open language models},
  author={Shao, Zhihong and Wang, Peiyi and Zhu, Qihao and Xu, Runxin and Song, Junxiao and Bi, Xiao and Zhang, Haowei and Zhang, Mingchuan and Li, YK and Wu, Yang and others},
  journal={arXiv preprint arXiv:2402.03300},
  year={2024}
}

@article{shinn2023reflexion,
  title={Reflexion: Language agents with verbal reinforcement learning},
  author={Shinn, Noah and Cassano, Federico and Gopinath, Ashwin and Narasimhan, Karthik and Yao, Shunyu},
  journal={Advances in neural information processing systems},
  volume={36},
  pages={8634--8652},
  year={2023}
}

@inproceedings{zhao2023expel,
  title={Expel: Llm agents are experiential learners},
  author={Zhao, Andrew and Huang, Daniel and Xu, Quentin and Lin, Matthieu and Liu, Yong-Jin and Huang, Gao},
  booktitle={Proceedings of the AAAI Conference on Artificial Intelligence},
  volume={38},
  number={17},
  pages={19632--19642},
  year={2024}
}

@article{nottingham2024sso,
  title={Skill set optimization: Reinforcing language model behavior via transferable skills},
  author={Nottingham, Kolby and Majumder, Bodhisattwa Prasad and Mishra, Bhavana Dalvi and Singh, Sameer and Clark, Peter and Fox, Roy},
  journal={arXiv preprint arXiv:2402.03244},
  year={2024}
}

@article{fu2024autoguide,
  title={Autoguide: Automated generation and selection of context-aware guidelines for large language model agents},
  author={Fu, Yao and Kim, Dong-Ki and Kim, Jaekyeom and Sohn, Sungryull and Logeswaran, Lajanugen and Bae, Kyunghoon and Lee, Honglak},
  journal={Advances in Neural Information Processing Systems},
  volume={37},
  pages={119919--119948},
  year={2024}
}

@article{chen2024automanual,
  title={Automanual: Constructing instruction manuals by llm agents via interactive environmental learning},
  author={Chen, Minghao and Li, Yihang and Yang, Yanting and Yu, Shiyu and Lin, Binbin and He, Xiaofei},
  journal={Advances in Neural Information Processing Systems},
  volume={37},
  pages={589--631},
  year={2024}
}

@article{wang2025awm,
  title={Agent workflow memory},
  author={Wang, Zora Zhiruo and Mao, Jiayuan and Fried, Daniel and Neubig, Graham},
  journal={arXiv preprint arXiv:2409.07429},
  year={2024}
}

@article{chhikara2025mem0,
  title={Mem0: Building production-ready ai agents with scalable long-term memory},
  author={Chhikara, Prateek and Khant, Dev and Aryan, Saket and Singh, Taranjeet and Yadav, Deshraj},
  journal={arXiv preprint arXiv:2504.19413},
  year={2025}
}

@inproceedings{fang2025memp,
  title={Memp: Exploring agent procedural memory},
  author={Fang, Runnan and Liang, Yuan and Wang, Xiaobin and Wu, Jialong and Qiao, Shuofei and Xie, Pengjun and Huang, Fei and Chen, Huajun and Zhang, Ningyu},
  booktitle={Findings of the Association for Computational Linguistics: ACL 2026},
  pages={17490--17502},
  year={2026}
}

@misc{wang2025sage,
      title={Reinforcement Learning for Self-Improving Agent with Skill Library}, 
      author={Jiongxiao Wang and Qiaojing Yan and Yawei Wang and Yijun Tian and Soumya Smruti Mishra and Zhichao Xu and Megha Gandhi and Panpan Xu and Lin Lee Cheong},
      year={2026},
      eprint={2512.17102},
      archivePrefix={arXiv},
      primaryClass={cs.AI},
      url={https://arxiv.org/abs/2512.17102}, 
}

@article{feng2025gigpo,
  title={Group-in-group policy optimization for llm agent training},
  author={Feng, Lang and Xue, Zhenghai and Liu, Tingcong and An, Bo},
  journal={Advances in Neural Information Processing Systems},
  volume={38},
  pages={46375--46408},
  year={2026}
}

@inproceedings{dong2025arpo,
  title={Agentic reinforced policy optimization},
  author={Dong, Guanting and Mao, Hangyu and Ma, Kai and Bao, Licheng and Chen, Yifei and Wang, Zhongyuan and Chen, Zhongxia and Du, Jiazhen and Wang, Huiyang and Zhang, Fuzheng and others},
  booktitle={International Conference on Learning Representations},
  volume={2026},
  pages={16981--17017},
  year={2026}
}

@article{skillrl2026,
  title={Skillrl: Evolving agents via recursive skill-augmented reinforcement learning},
  author={Xia, Peng and Chen, Jianwen and Wang, Hanyang and Liu, Jiaqi and Zeng, Kaide and Wang, Yu and Han, Siwei and Zhou, Yiyang and Zhao, Xujiang and Chen, Haifeng and others},
  journal={arXiv preprint arXiv:2602.08234},
  year={2026}
}

@article{zhu2026skill05,
  title={Skill0. 5: Joint Skill Internalization and Utilization for Out-of-Distribution Generalization in Agentic Reinforcement Learning},
  author={Zhu, Jiapeng and Yu, Jianxiang and Zhao, Yibo and Han, Chengcheng and Gu, Qi and Cai, Xunliang and Li, Xiang and Qian, Weining},
  journal={arXiv preprint arXiv:2605.28424},
  year={2026}
}

@article{lu2026skill0,
  title={Skill0: In-context agentic reinforcement learning for skill internalization},
  author={Lu, Zhengxi and Yao, Zhiyuan and Wu, Jinyang and Han, Chengcheng and Gu, Qi and Cai, Xunliang and Lu, Weiming and Xiao, Jun and Zhuang, Yueting and Shen, Yongliang},
  journal={arXiv preprint arXiv:2604.02268},
  year={2026}
}

@article{wang2026skillsd,
  title={Skill-sd: Skill-conditioned self-distillation for multi-turn llm agents},
  author={Wang, Hao and Wang, Guozhi and Xiao, Han and Zhou, Yufeng and Pan, Yue and Wang, Jichao and Xu, Ke and Wen, Yafei and Ruan, Xiaohu and Chen, Xiaoxin and others},
  journal={arXiv preprint arXiv:2604.10674},
  year={2026}
}

@article{seed2026,
  title={SEED: Self-Evolving On-Policy Distillation for Agentic Reinforcement Learning},
  author={Wu, Jinyang and Yang, Shuo and Lu, Zhengxi and Zhang, Fan and Shen, Yuhao and Feng, Lang and Luo, Haoran and Lian, Zheng and Zhang, Shuai and Wen, Zhengqi and others},
  journal={arXiv preprint arXiv:2607.14777},
  year={2026}
}

@article{shi2026skill1,
  title={Skill1: Unified evolution of skill-augmented agents via reinforcement learning},
  author={Shi, Yaorui and Chen, Yuxin and Lu, Zhengxi and Miao, Yuchun and Liu, Shugui and Gu, Qi and Cai, Xunliang and Wang, Xiang and Zhang, An},
  journal={arXiv preprint arXiv:2605.06130},
  year={2026}
}

@article{fan2026evolvingrl,
  title={Evolving-RL: End-to-End Optimization of Experience-Driven Self-Evolving Capability within Agents},
  author={Fan, Zhiyuan and Jin, Wenwei and Zhang, Feng and Li, Bin and Dong, Yihong and Hu, Yao and Li, Jiawei},
  journal={arXiv preprint arXiv:2605.10663},
  year={2026}
}

@article{zhou2026inspo,
  title={Agentic Policy Optimization via Instruction-Policy Co-Evolution},
  author={Zhou, Han and Wan, Xingchen and Vuli{\'c}, Ivan and Korhonen, Anna},
  journal={arXiv preprint arXiv:2512.01945},
  year={2025}
}

@article{zhang2026sapo,
  title={Co-Evolving Skill Generation and Policy Optimization},
  author={Zhang, Zhiwei and Lin, Yudi and Kuang, Nikki Lijing and Wu, Linlin and Li, Xiaomin and Liu, Songtao and Ma, Fenglong},
  journal={arXiv preprint arXiv:2606.08755},
  year={2026}
}

@article{shen2026slim,
  title={Dynamic skill lifecycle management for agentic reinforcement learning},
  author={Shen, Junhao and Zhang, Teng and Zhao, Xiaoyan and Cheng, Hong},
  journal={arXiv preprint arXiv:2605.10923},
  year={2026}
}

@article{he2026reskill,
  title={ReSkill: Reconciling Skill Creation with Policy Optimization in Agentic RL},
  author={He, Zelin and Lin, Haotian and Han, Boran and Zhu, Wei and Fang, Haoyang and Wang, Bernie and Zhu, Xuan and Li, Runze and Reimherr, Matthew},
  journal={arXiv preprint arXiv:2606.01619},
  year={2026}
}

@article{dong2026harmfulskills,
  title={Agent Skills Can Be Harmful: An Empirical Study of Skill-Induced Failures in LLM Agents},
  author={Dong, Gen and Gao, Yanjie and Li, Liqun and Xu, Tianyin and Hua, Yu and Yang, Fan},
  journal={arXiv preprint arXiv:2608.11888},
  year={2026}
}

@article{gao2026arealsea,
  title={From self-evolving synthetic data to verifiable-reward rl: Post-training multi-turn interactive tool-using agents},
  author={Gao, Jiaxuan and Chen, Jiaao and He, Chuyi and Xu, Shusheng and Jin, Di and Wu, Yi},
  journal={arXiv preprint arXiv:2601.22607},
  year={2026}
}

@article{shridhar2021alfworld,
  title={Alfworld: Aligning text and embodied environments for interactive learning},
  author={Shridhar, Mohit and Yuan, Xingdi and C{\^o}t{\'e}, Marc-Alexandre and Bisk, Yonatan and Trischler, Adam and Hausknecht, Matthew},
  journal={arXiv preprint arXiv:2010.03768},
  year={2020}
}

@article{yao2022webshop,
  title={Webshop: Towards scalable real-world web interaction with grounded language agents},
  author={Yao, Shunyu and Chen, Howard and Yang, John and Narasimhan, Karthik},
  journal={Advances in Neural Information Processing Systems},
  volume={35},
  pages={20744--20757},
  year={2022}
}

@misc{wang2022scienceworld,
      title={ScienceWorld: Is your Agent Smarter than a 5th Grader?}, 
      author={Ruoyao Wang and Peter Jansen and Marc-Alexandre Côté and Prithviraj Ammanabrolu},
      year={2022},
      eprint={2203.07540},
      archivePrefix={arXiv},
      primaryClass={cs.CL},
      url={https://arxiv.org/abs/2203.07540}, 
}

@misc{yao2024taubench,
      title={$\tau$-bench: A Benchmark for Tool-Agent-User Interaction in Real-World Domains}, 
      author={Shunyu Yao and Noah Shinn and Pedram Razavi and Karthik Narasimhan},
      year={2024},
      eprint={2406.12045},
      archivePrefix={arXiv},
      primaryClass={cs.AI},
      url={https://arxiv.org/abs/2406.12045}, 
}
\bibliographystyle{iclr2027_conference}

\appendix

\clearpage

\section{Implementation Details}
\label{app:implementation}

\subsection{Benchmark and Data Details}
\label{app:benchmark-details}

\paragraph{ALFWorld.}
ALFWorld contains six household task categories: Pick, Heat, Cool, Clean,
Look, and Pick2. We train on the full 3,553-task split, comprising 790 Pick,
459 Heat, 533 Cool, 650 Clean, 308 Look, and 813 Pick2 tasks, and evaluate
success rate on the 134-task valid-unseen split.

\paragraph{WebShop.}
Following the category-generalization protocol of Skill0.5
\citep{zhu2026skill05}, we partition the 12,087 human-annotated WebShop goals
into seven domains using keyword-based classification of goal instructions.
Apparel, Electronics, Footwear, and Other are used for training and ID
evaluation; Accessories, Beauty \& Health, and Home Decor are held out for OOD
evaluation. Because Other initially accounts for over 60\% of the goals, we
apply farthest-point sampling over sentence embeddings to balance its size.

The resulting training split contains 3,320 ID goals: 776 Apparel, 938
Electronics, 606 Footwear, and 1,000 Other goals, with Other downsampled from
approximately 6,600. The ID validation split contains 454 goals: 113 Apparel,
152 Electronics, 89 Footwear, and 100 Other, with Other downsampled from
approximately 940. The OOD validation split contains 207 goals: 80
Accessories, 54 Beauty \& Health, and 73 Home Decor. Training uses original
goal indices 1,500 onward, while the validation pool combines the original
test indices 0--499 and development indices 500--1,499.

\paragraph{ScienceWorld.}
Following ReSkill~\citep{he2026reskill}, we use the electricity split, which
contains two task types: identifying power components and testing
conductivity. The action space contains approximately 26 verb templates, and
the interaction horizon is at most 50 steps. Training uses 460 variations: 10
\texttt{power-component} and 450 \texttt{test-conductivity} variations under
the \texttt{easy} simplification. Test-ID uses 230 development episodes with
held-out variations of the same task types; Test-OOD uses 155 episodes from
related variants involving unknown-material conductivity and renewable
energy. We report ID and OOD success rates.

\paragraph{$\tau^2$-Bench.}
$\tau^2$-Bench evaluates long-horizon dialogue, tool execution, and policy
compliance in Airline, Retail, and Telecom. Unlike the other three benchmarks,
all RL-based baselines and \method are trained from the same SFT checkpoint.
The user role is simulated by DeepSeek-V4-Flash.
Both the SFT and RL data are derived from the synthetic data released by
AReaL-SEA~\citep{gao2026arealsea}. We estimate task pass rates with the SFT
model and remove all-success and all-failure groups, which provide no
group-relative learning signal. The resulting RL pool contains 300 entries per
domain, or 900 in total.

\subsection{Benchmark Configurations}

Table~\ref{tab:app-training-config} summarizes the configurations used during
training.
ALFWorld, WebShop, and ScienceWorld are initialized from the base Qwen3-8B
model, whereas all RL-based methods on $\tau^2$-Bench use the same
domain-adapted SFT checkpoint.

\begin{table}[H]
\centering
\caption{Training configurations for the reported \method models.}
\label{tab:app-training-config}
\begingroup
\small
\setlength{\tabcolsep}{4pt}
\renewcommand{\arraystretch}{1.08}
\begin{tabular*}{\linewidth}{@{\extracolsep{\fill}}lcccc@{}}
\toprule
Parameter & ALFWorld & WebShop & ScienceWorld & $\tau^2$-Bench \\
\midrule
\multicolumn{5}{@{}l}{\emph{Reinforcement learning}} \\
Model family & Qwen3-8B & Qwen3-8B & Qwen3-8B & Qwen3-8B \\
Training steps & 100 & 200 & 160 & 200 \\
Optimizer & Adam & Adam & Adam & Adam \\
Learning rate (constant) & $10^{-6}$ & $10^{-6}$ & $10^{-6}$ & $10^{-6}$ \\
Adam $(\beta_1,\beta_2)$ & $(0.9,0.98)$ & $(0.9,0.98)$ & $(0.9,0.98)$ & $(0.9,0.999)$ \\
Query batch size & 16 & 16 & 16 & 32 \\
Configured global batch size & 128 & 128 & 128 & 32 \\
Executor group size & 8 & 8 & 8 & 8 \\
PPO clip $\epsilon$ (lower / upper) & 0.20 / 0.28 & 0.20 / 0.28 & 0.20 / 0.28 & 0.20 / 0.28 \\
KL penalty & 0 & 0 & 0 & 0 \\
Executor entropy coefficient & 0 & 0 & 0 & 0 \\
Executor temperature & 1.0 & 1.0 & 1.0 & 0.7 \\
Executor top-$p$ & 0.85 & 0.85 & 0.85 & 1.0 \\
Maximum interaction steps & 50 & 15 & 50 & 160 \\
Maximum Executor tokens / turn & 512 & 512 & 384 & 1,024 \\
Executor sequence token budget & 32,768 & 16,384 & 32,768 & 32,768 \\
\midrule
\multicolumn{5}{@{}l}{\emph{Skill generation and selection}} \\
Utility variant & Group-based & Anchor-based & Group-based & Group-based \\
$N$ Scouts / $M$ candidates / $L$ Post-Scouts
& 4 / 4 / 4 & 4 / 4 / 4 & 4 / 4 / 4 & 4 / 4 / 4 \\
Maximum Generator tokens & 900 & 256 & 900 & 700 \\
Generator-loss weight $\lambda_{\mathrm{gen}}$ & 0.1 & 0.1 & 0.1 & 0.1 \\
Generator entropy coefficient & 0.002 & 0 & 0.002 & 0.002 \\
Injection threshold & $U>0$ & $U>0$ & $U>0$ & $U>0$ \\
Test-time external skill & None & None & None & None \\
\bottomrule
\end{tabular*}
\endgroup
\end{table}

\subsection{End-to-End Training Procedure}

Algorithm~\ref{alg:skillscaffold} summarizes one complete \method training
update. Environment rewards supervise the Executor, whereas detached
policy-relative utility supervises the Generator. Both objectives update
the same policy parameters.

\begin{algorithm}[t]
\caption{\method training with policy-relative skill utility}
\label{alg:skillscaffold}
\begin{algorithmic}[1]
\Require Policy $\pi_\theta$, task distribution $\mathcal{D}$, tool engine
$\mathcal{T}$, utility variant $v\in\{\mathrm{grp},\mathrm{anc}\}$
\Require $N$ Scouts, $M$ candidate skills, $L$ Post-Scout rollouts
\For{each training update}
    \State Sample $x\sim\mathcal{D}$
    \State Collect skill-free Scouts
    $\{(\tau_i^0,y_i)\}_{i=1}^{N}$ with
    $\pi_{\theta}^{\mathrm{exec}}$
    \State Construct evidence $E_x^v$ from Scout trajectories and outcomes
    \If{$v=\mathrm{anc}$ and no Scout succeeds}
        \State Set $\mathcal{C}_x\gets\varnothing$ and
        $s_x^*\gets\varnothing$
    \Else
        \State Generate $\mathcal{C}_x=\{s_j\}_{j=1}^{M}$ using
        $\pi_{\theta}^{\mathrm{gen}}(\cdot\mid E_x^v)$
        \For{each format-valid candidate $s_j$}
            \State Score the fixed Scout tokens under $s_j$ and the skill-free
            context
            \State Compute utility $U_j\gets U^v(s_j;x)$
        \EndFor
        \State Let $\mathcal{V}_x$ contain candidates with valid format and
        finite utility
        \If{$\mathcal{V}_x=\varnothing$}
            \State $s_x^*\gets\varnothing$
        \Else
            \State $j^*\gets\arg\max_{j\in\mathcal{V}_x} U_j$
            \State $s_x^*\gets s_{j^*}$ if $U_{j^*}>0$; otherwise
            $s_x^*\gets\varnothing$
        \EndIf
    \EndIf
    \State Collect $L$ fresh Post-Scout trajectories conditioned on $s_x^*$
    \State Form the Executor GRPO group from the $N+L$ trajectories using
    environment rewards
    \State Form Generator rewards using Equation~\ref{eq:generator-reward}
    when a complete trainable group is available
    \State Update shared parameters with
    $J_{\mathrm{GRPO}}^{\mathrm{exec}}
    +\lambda_{\mathrm{gen}}J_{\mathrm{GRPO}}^{\mathrm{gen}}
    +\eta_{\mathrm{gen}}\mathcal{H}_{\mathrm{gen}}$; omit Generator terms
    when unavailable
\EndFor
\State \textbf{return} skill-free policy
$\pi_{\theta^*}^{\mathrm{exec}}(\cdot\mid x,\varnothing;\mathcal{T})$
\end{algorithmic}
\end{algorithm}

\subsection{Evidence Construction and Candidate Generation}

For group-based training, the evidence record contains the task, all $N$
Scout trajectories, and their verifier outcomes. All-success groups emphasize
reusable successful decisions, all-failure groups request conditional
corrections without inventing an unobserved solution, and mixed groups contrast
successful and failed decisions. For anchor-based training, the first verified
successful Scout in sampling order is used as the evidence anchor. If no Scout
succeeds, candidate generation and Generator optimization are skipped for that
query, and Executor execution remains skill-free.

Each evidence record produces $M$ independently sampled candidate skills.
Candidates must satisfy the prescribed card format and express reusable
decision guidance rather than copy task-specific answers. Formatting validity
is checked separately from utility availability: an invalid candidate is never
selected, while an incomplete Generator group is omitted according to the
rules in Section~\ref{sec:joint-optimization}.

\subsection{Diagnostic Evaluation Protocols}
\label{app:diagnostic-protocols}

\paragraph{Skill-utility validation.}
The comparison in Section~\ref{sec:skill-credit-validation} applies every
injection rule to the same fixed query--candidate sets. Candidate utilities are
computed from the fixed Scout trajectories, while separate downstream rollouts
are reserved for evaluation. The utility-quintile analysis groups
candidate--query pairs by the estimated utility and reports their independently
measured gain over skill-free execution.

\paragraph{Skill-free strategy acquisition.}
For each benchmark, we form a fixed cohort at an early checkpoint from training
queries that benefit from skill-assisted exploration. Cohort membership is
frozen before later evaluation. The same queries are then reevaluated at a
later checkpoint with the skill pathway disabled, and recovery is computed as
the later-minus-early skill-free success rate. Confidence intervals use a
query-level bootstrap.


\section{Additional Experimental Results}
\label{app:additional-results}

\subsection{Full \texorpdfstring{$\tau^2$-Bench Pass@$k$}{Tau2-Bench Pass@k} Results}

Table~\ref{tab:tau2-full} expands the main Pass@1 comparison with Pass@4 and
Pass@8 for every domain. The average columns are macro averages over Airline,
Retail, and Telecom.

\begin{table*}[h]
\centering
\caption{Full $\tau^2$-Bench Pass@$k$ results (\%) with Qwen3-8B. Average columns
are macro averages over Airline, Retail, and Telecom. Teal arrows indicate
absolute gains in percentage points over the strongest competing result.}
\label{tab:tau2-full}
\begingroup
\scriptsize
\setlength{\tabcolsep}{2.2pt}

\begin{tabular}{lrrrrrrrrrrrr}
\toprule
& \multicolumn{3}{c}{Airline}
& \multicolumn{3}{c}{Retail}
& \multicolumn{3}{c}{Telecom}
& \multicolumn{3}{c}{Average} \\
\cmidrule(lr){2-4}\cmidrule(lr){5-7}\cmidrule(lr){8-10}\cmidrule(lr){11-13}

Method
& P@1 & P@4 & P@8
& P@1 & P@4 & P@8
& P@1 & P@4 & P@8
& P@1 & P@4 & P@8 \\
\midrule

ReAct
& 17.8 & 34.0 & 46.0
& 35.5 & 61.7 & 70.2
& 7.6 & 23.1 & 32.5
& \cellcolor{avgbg}20.3 & \cellcolor{avgbg}39.6 & \cellcolor{avgbg}49.5 \\

SFT
& 41.0 & 62.1 & \second{74.0}
& 40.1 & 71.3 & 84.2
& 56.6 & 91.8 & 97.4
& \cellcolor{avgbg}45.9 & \cellcolor{avgbg}75.1 & \cellcolor{avgbg}85.2 \\

GRPO
& \second{45.8} & 63.8 & 70.0
& 57.2 & \second{85.4} & \best{92.1}
& \second{77.0} & \second{97.2} & \second{99.1}
& \cellcolor{avgbg}\second{60.0} & \cellcolor{avgbg}\second{82.1} & \cellcolor{avgbg}\second{87.1} \\

SkillRL
& \best{46.0} & \second{64.2} & 70.0
& \second{57.9} & 83.6 & 90.4
& 75.7 & 95.3 & \second{99.1}
& \cellcolor{avgbg}59.9 & \cellcolor{avgbg}81.0 & \cellcolor{avgbg}86.5 \\

\ours
& \best{46.0} & \best{70.3\gain{6.1}} & \best{78.0\gain{4.0}}
& \best{60.8\gain{2.9}} & \best{85.6\gain{0.2}} & \second{91.2}
& \best{83.3\gain{6.3}} & \best{98.5\gain{1.3}} & \best{100.0\gain{0.9}}
& \best{63.4\gain{3.4}} & \best{84.8\gain{2.7}} & \best{89.7\gain{2.6}} \\

\bottomrule
\end{tabular}
\endgroup
\end{table*}

\subsection{Training Dynamics Across Benchmarks}

\begin{figure*}[t]
    \centering
    \includegraphics[width=\textwidth]
    {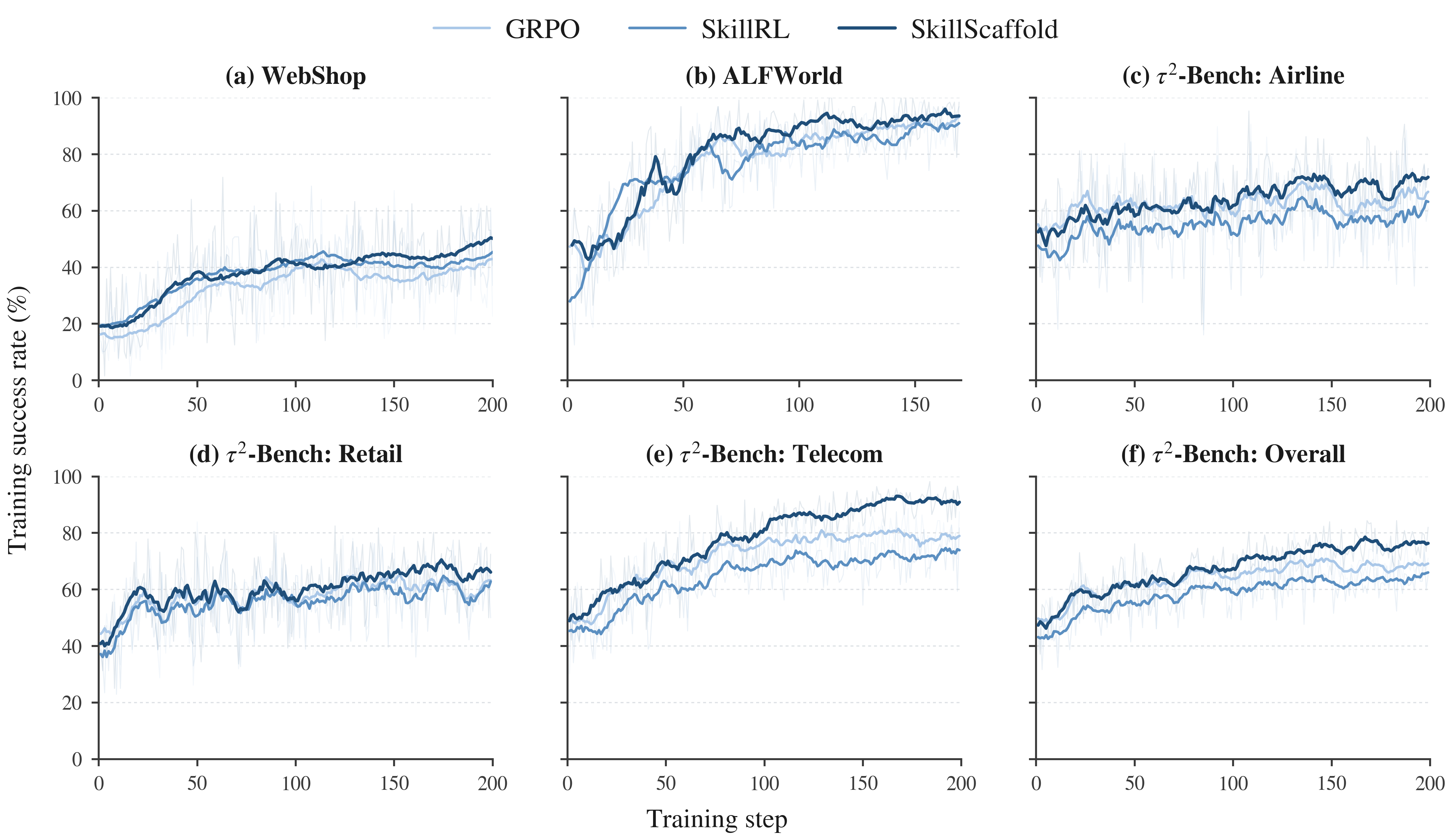}
    \caption{\textbf{Training success-rate curves across benchmarks.}
    GRPO, SkillRL, and \method on (a) WebShop, (b) ALFWorld,
    (c) $\tau^2$-Bench Airline, (d) Retail, (e) Telecom, and (f) the
    $\tau^2$-Bench macro average. Dark curves show smoothed trends and faint
    curves show raw measurements.}
    \label{fig:training-dynamics-full}
\end{figure*}

Figure~\ref{fig:training-dynamics-full} reports training curves for WebShop,
ALFWorld, and the three individual $\tau^2$-Bench domains. The largest
late-training separation appears on Telecom, and the Airline and Retail curves
also retain advantages at later checkpoints.

\subsection{Skill Injection Rates Across Benchmarks}

\begin{figure*}[t]
    \centering
    \begin{minipage}[t]{0.323\textwidth}
        \centering
        \includegraphics[width=\linewidth]
        {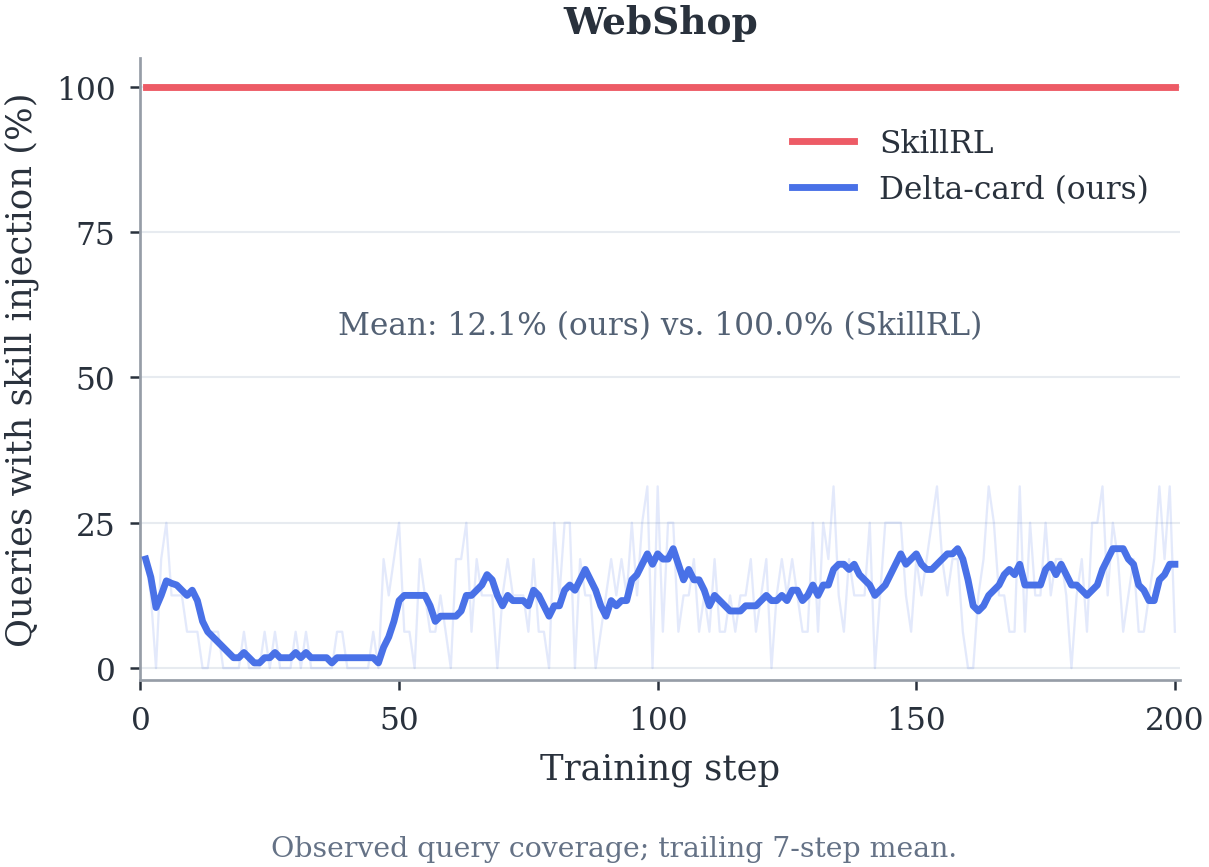}
        \small (a) WebShop
    \end{minipage}\hfill
    \begin{minipage}[t]{0.323\textwidth}
        \centering
        \includegraphics[width=\linewidth]
        {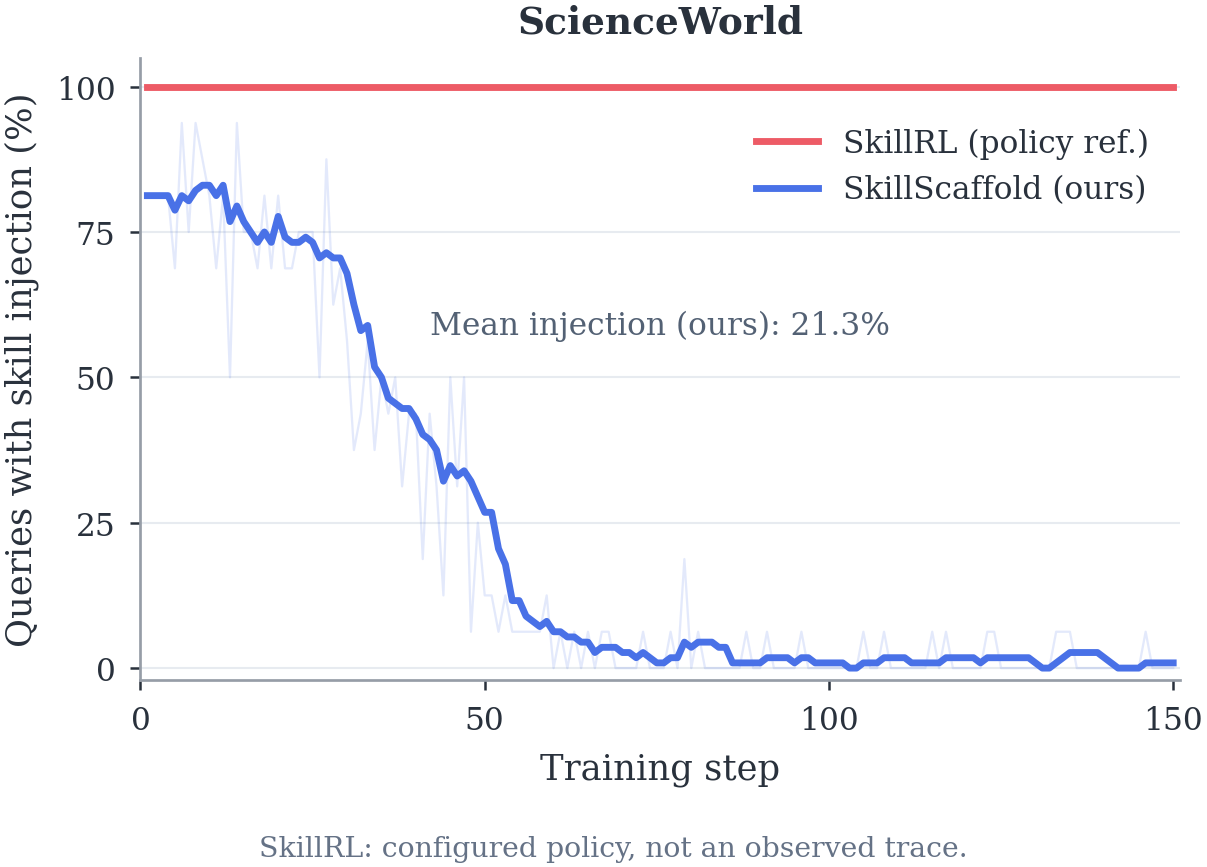}
        \small (b) ScienceWorld
    \end{minipage}\hfill
    \begin{minipage}[t]{0.323\textwidth}
        \centering
        \includegraphics[width=\linewidth]
        {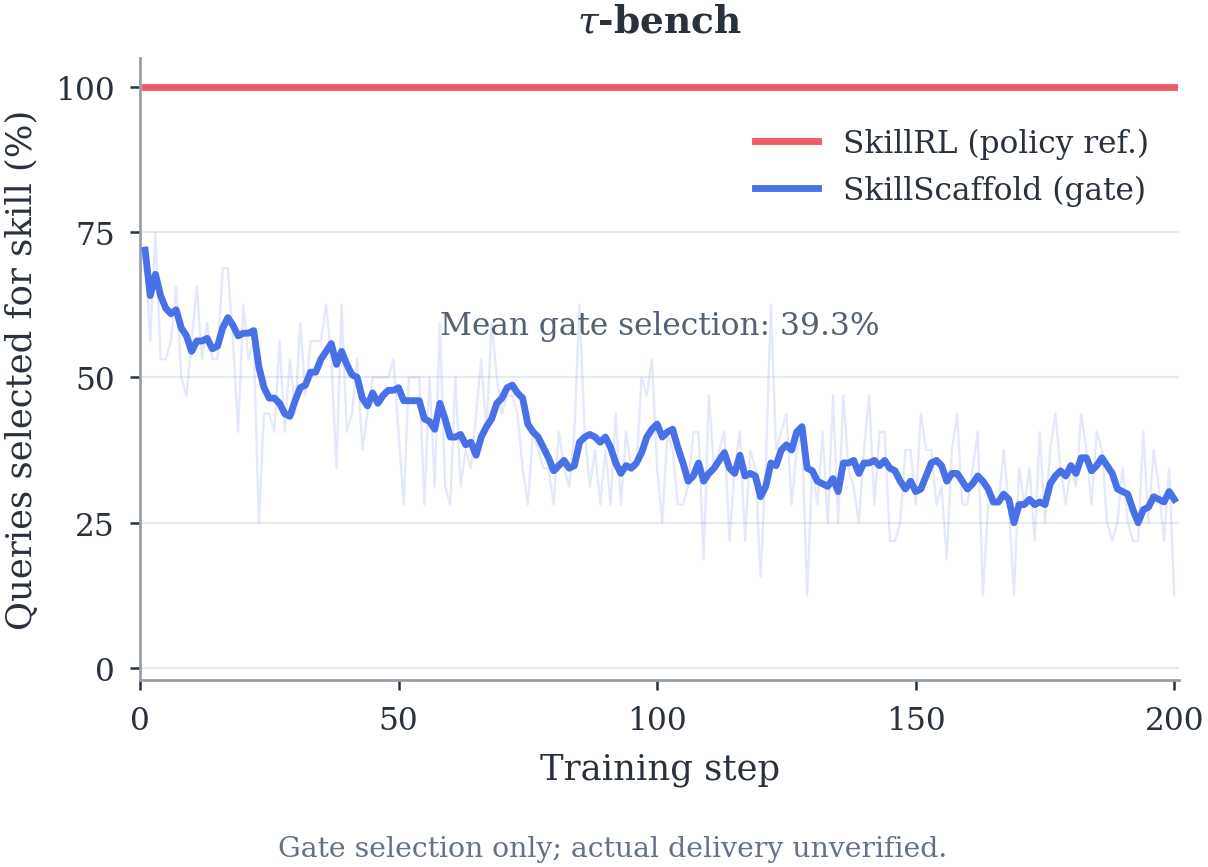}
        \small (c) $\tau^2$-Bench
    \end{minipage}
    \caption{\textbf{Training-time skill injection and selection rates.}
    Panels (a)--(b) report skill injection on WebShop and ScienceWorld;
    panel (c) reports positive-utility selection on $\tau^2$-Bench.}
    \label{fig:additional-injection-rates}
\end{figure*}

Figure~\ref{fig:additional-injection-rates} reports how frequently the
training-time scaffold is activated beyond ALFWorld. The always-on SkillRL
reference supplies a skill to every query, whereas \method invokes its
scaffold selectively. The mean observed injection rate is 12.1\% on WebShop
and 21.3\% on ScienceWorld. On $\tau^2$-Bench, the positive-utility gate selects
a skill for 39.3\% of queries on average.

\subsection{Cross-Benchmark Post-Scout Exploration Dynamics}

\begin{figure*}[t]
    \centering
    \includegraphics[width=0.98\textwidth]
    {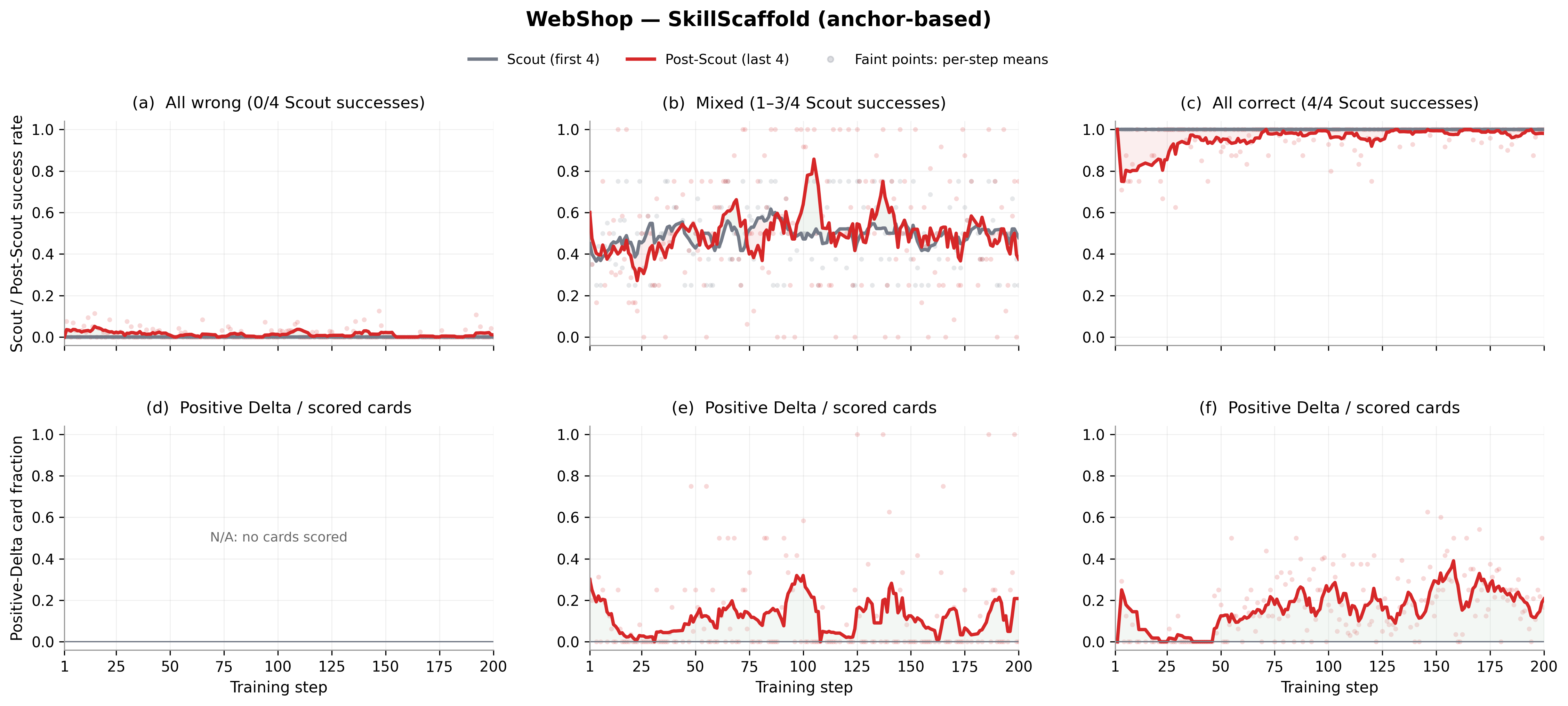}
    \caption{\textbf{Selective anchor-guided exploration on WebShop.}
    Using a verified successful Scout as evidence, the anchor-based
    variant selectively guides mixed groups and maintains strong Post-Scout
    performance as the Scout policy approaches the all-correct regime.}
    \label{fig:webshop-anchor-postscout}
\end{figure*}

\begin{figure*}[t]
    \centering
    \includegraphics[width=0.98\textwidth]
    {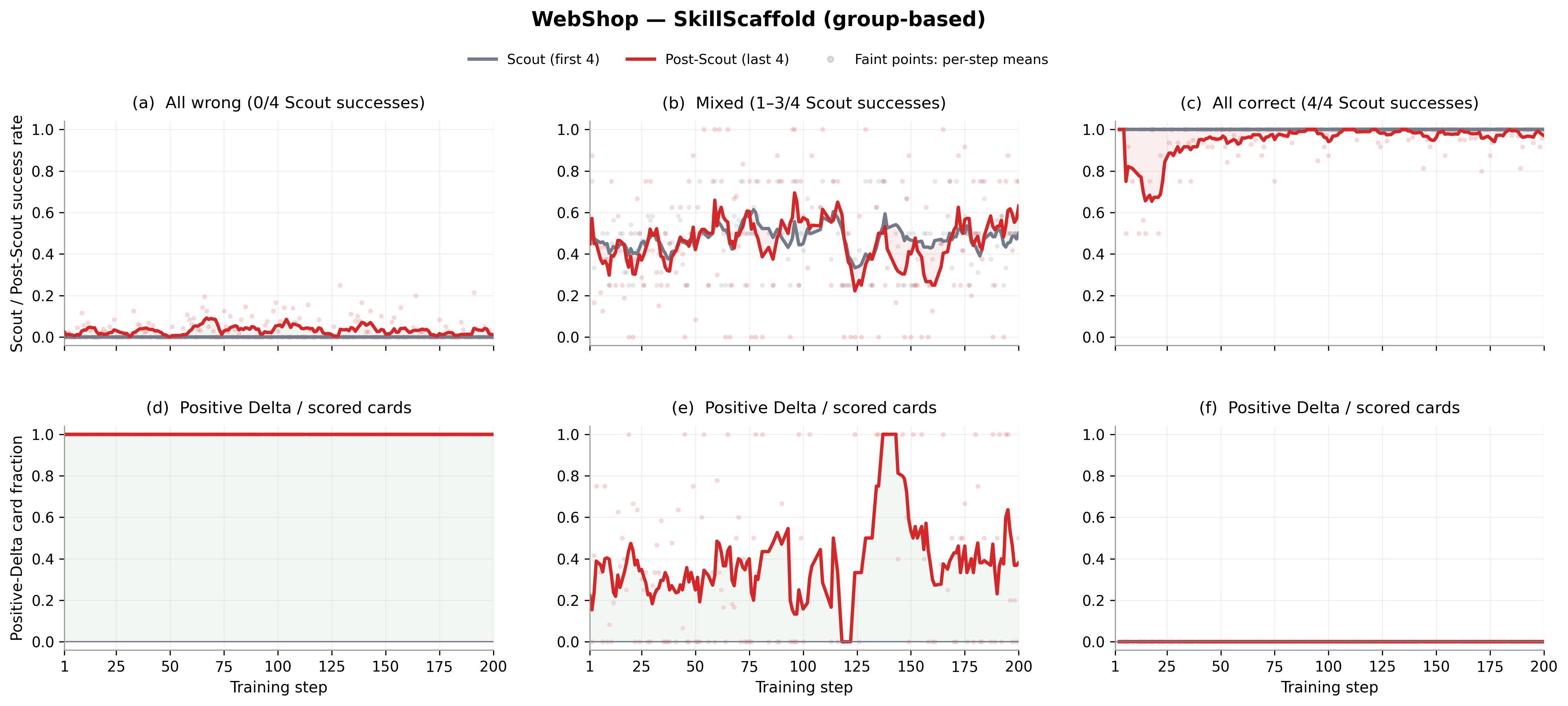}
    \caption{\textbf{Corrective group-guided exploration on WebShop.}
    Outcome-labeled Scout groups support corrective skill generation
    and produce persistent nonzero Post-Scout success for initially all-wrong
    queries. Skill activation concentrates on all-wrong and mixed groups and
    naturally diminishes once all Scouts succeed.}
    \label{fig:webshop-group-postscout}
\end{figure*}

\begin{figure*}[t]
    \centering
    \includegraphics[width=0.98\textwidth]
    {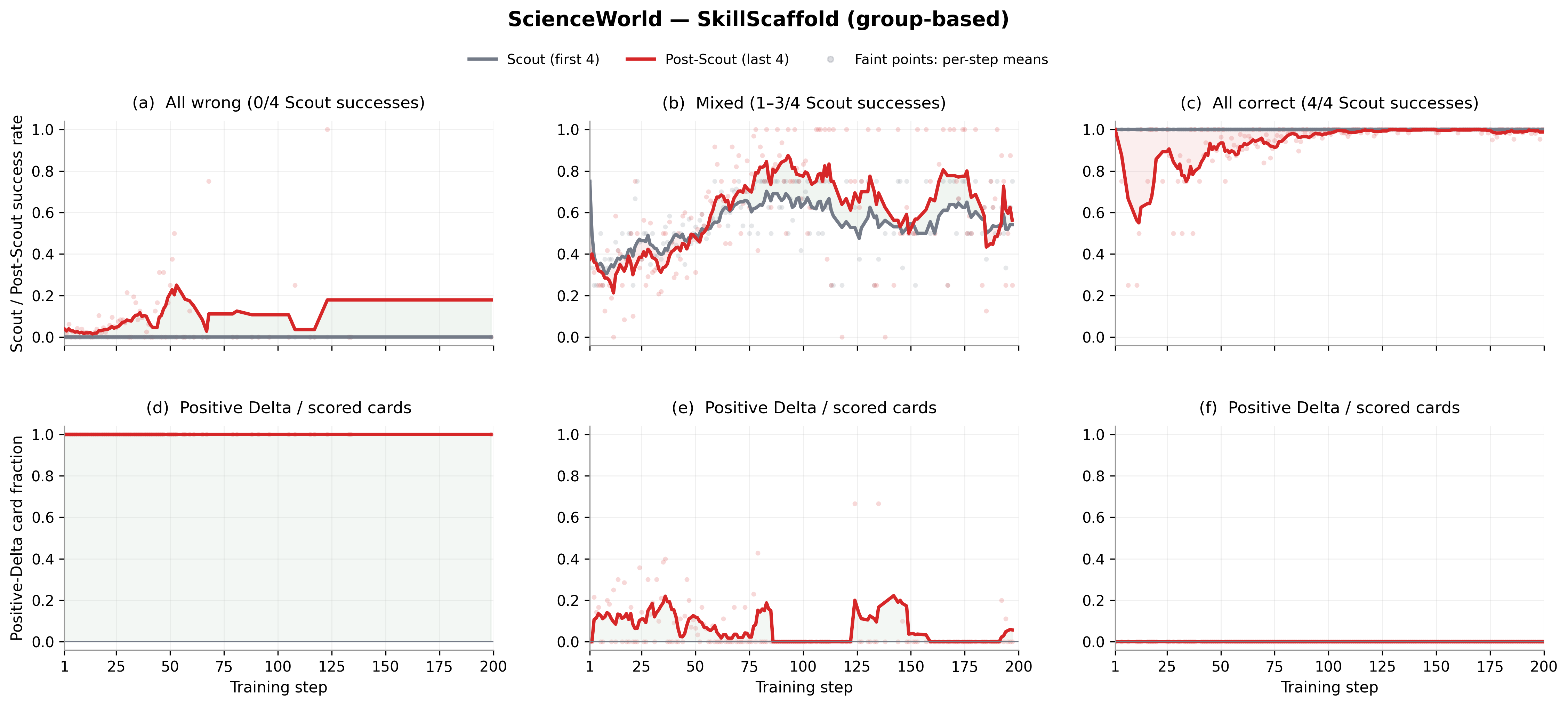}
    \caption{\textbf{Competence-adaptive exploration on ScienceWorld.}
    Group-based scaffolding creates successful trajectories for initially all-wrong groups
    and improves mixed-group performance during later training. Positive-utility
    candidates become less frequent as the Scout policy improves, reflecting
    competence-adaptive skill use.}
    \label{fig:scienceworld-postscout}
\end{figure*}

\begin{figure*}[t]
    \centering
    \includegraphics[width=0.98\textwidth]
    {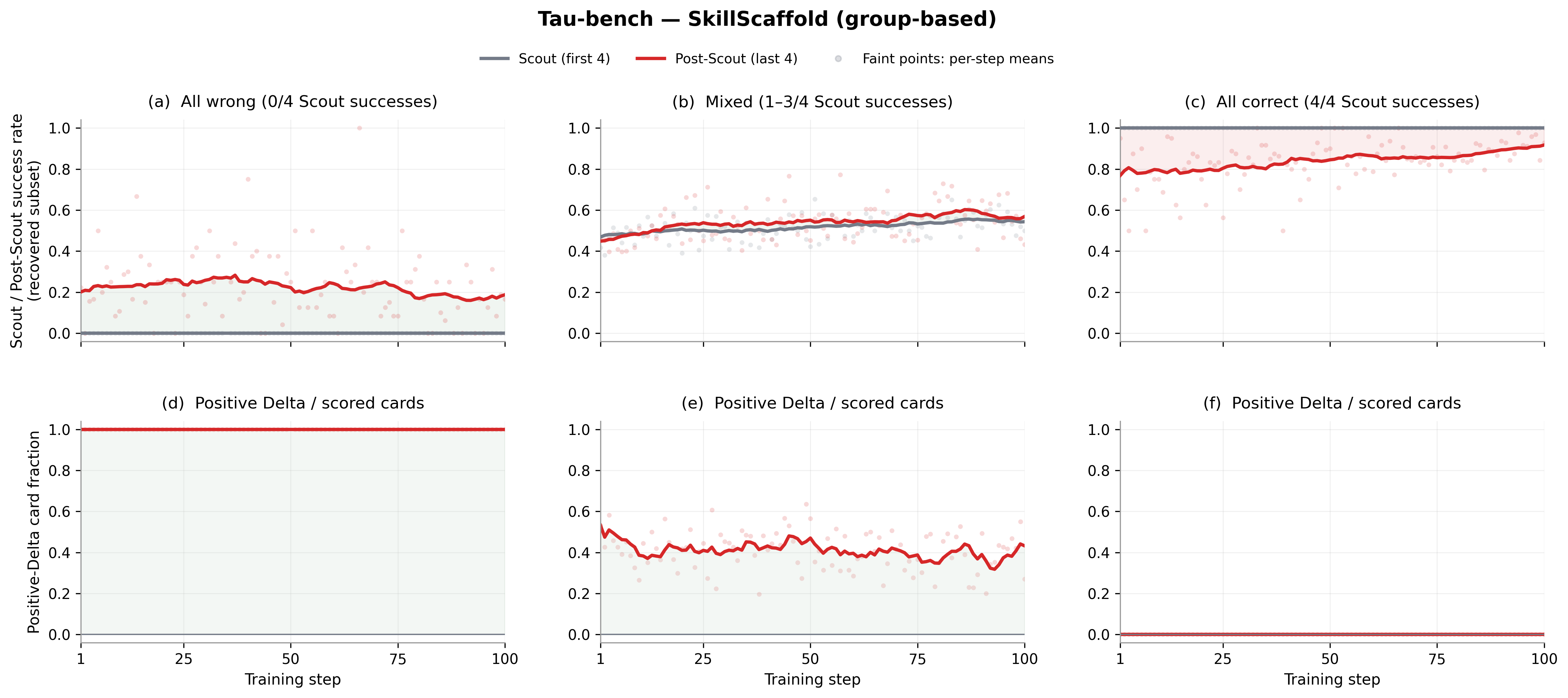}
    \caption{\textbf{Competence-adaptive exploration on $\tau^2$-Bench.}
    Post-Scout rollouts recover successful trajectories from initially all-wrong groups
    and improve performance across much of the mixed regime. Positive-utility
    candidates are frequent for all-wrong groups, selective for mixed groups,
    and unnecessary once all Scouts succeed.}
    \label{fig:tau-postscout}
\end{figure*}

Figures~\ref{fig:webshop-anchor-postscout}--\ref{fig:tau-postscout}
test whether the selective-scaffolding behavior observed on ALFWorld also
generalizes to the other benchmarks. Queries are grouped by the outcomes of
four skill-free Scouts. The upper row compares Scout and fresh Post-Scout
rewards, while the lower row reports the positive-utility candidate rate.

Across benchmarks, \method concentrates positive-utility skills on all-wrong
and mixed groups, where Post-Scout rollouts recover additional successful
behavior. Skill use declines as the Scout policy becomes competent and is rare
once all Scouts succeed, focusing external guidance on queries where it
improves exploration.

\section{Theoretical Analysis of Skill Utility for Credit Assignment}
\label{app:theory}

This section analyzes the proposed likelihood-based skill utility as a
policy-relative surrogate. We derive its log-likelihood-ratio interpretation,
ranking and abstention robustness, a calibrated downstream-return bound, and
its interaction and computation complexity.

\subsection{Log-Likelihood-Ratio Interpretation}

For the Executor-token index set $\mathcal{I}(\tau)$, define the conditional
trajectory likelihood
\begin{equation}
p_{\theta}(\tau\mid x,c)
=
\prod_{t\in\mathcal{I}(\tau)}
\pi_{\theta}^{\mathrm{exec}}
(u_t\mid x,c,h_t^\tau).
\label{eq:app-trajectory-probability}
\end{equation}
Equation~\ref{eq:trajectory-likelihood} then satisfies the exact identity
\begin{equation}
d(s,\tau)
=
\frac{1}{|\mathcal{I}(\tau)|}
\log
\frac{p_{\theta}(\tau\mid x,s)}
     {p_{\theta}(\tau\mid x,\varnothing)}.
\label{eq:app-log-ratio}
\end{equation}
A positive $d(s,\tau)$ means that conditioning on $s$ increases the geometric
mean token likelihood assigned to the observed Executor behavior.

\paragraph{Proposition 1 (balanced signed evidence).}
For a mixed Scout group, group-based utility can be written as
\begin{equation}
U^{\mathrm{grp}}(s;x)
=
\frac{1}{2}
\mathbb{E}_{i\sim\mathrm{Unif}(\mathcal{S}_x)}[d(s,\tau_i^0)]
-
\frac{1}{2}
\mathbb{E}_{i\sim\mathrm{Unif}(\mathcal{F}_x)}[d(s,\tau_i^0)].
\label{eq:app-balanced-utility}
\end{equation}
Consequently, successful and failed evidence each receive total weight
$1/2$, independent of their relative frequency in the Scout group.

\emph{Proof.}
Substituting the uniform expectations into
Equation~\ref{eq:app-balanced-utility} gives the mixed branch of
Equation~\ref{eq:scaffold-utility}. The coefficients sum to $1/2$ within each
outcome class, so changing the class counts does not change their total
weights. \hfill$\square$

This balancing keeps successful and failed evidence equally influential,
independent of their relative frequencies in the sampled Scout group.

\subsection{Selection Robustness under Utility Error}

Let $U_j$ denote the population proxy utility of candidate $j$ under the
current query and policy, and let $\widehat U_j$ be the utility estimated
from the sampled Scout group. Assume a uniform error bound
\begin{equation}
|\widehat U_j-U_j|\leq\varepsilon
\qquad\text{for every eligible candidate }j.
\label{eq:app-uniform-error}
\end{equation}

\paragraph{Proposition 2 (ranking and abstention bounds).}
Let $\widehat j=\arg\max_j\widehat U_j$ and
$j^*=\arg\max_j U_j$. Then
\begin{equation}
U_{\widehat j}
\geq
U_{j^*}-2\varepsilon.
\label{eq:app-selection-regret}
\end{equation}
Moreover, if the positive gate selects $\widehat j$, then
$U_{\widehat j}>-\varepsilon$; if the gate abstains, then
$\max_j U_j\leq\varepsilon$.

\emph{Proof.}
By optimality of $\widehat j$ under the estimated utilities,
\begin{equation}
U_{\widehat j}
\geq
\widehat U_{\widehat j}-\varepsilon
\geq
\widehat U_{j^*}-\varepsilon
\geq
U_{j^*}-2\varepsilon.
\end{equation}
If $\widehat U_{\widehat j}>0$, the error bound gives
$U_{\widehat j}>-\varepsilon$. If every estimated utility is nonpositive,
then $U_j\leq\widehat U_j+\varepsilon\leq\varepsilon$ for every
candidate. \hfill$\square$

The proposition formalizes two roles of utility in skill credit assignment. Ranking incurs at most
$2\varepsilon$ proxy regret, while abstention avoids overlooking any candidate
whose proxy advantage exceeds the estimation-error scale.

\subsection{Connection to Downstream Return}

Let
\begin{equation}
G(s;x)
=
\mathbb{E}[R(\tau;x)\mid x,s]
-
\mathbb{E}[R(\tau;x)\mid x,\varnothing]
\label{eq:app-true-gain}
\end{equation}
denote the true downstream gain. The following local calibration condition
connects policy-relative utility to this environment-level quantity.

\paragraph{Assumption 1 (local surrogate calibration).}
For eligible candidates under the current query and policy, suppose constants
$a>0$ and $b\geq0$ exist such that
\begin{equation}
G(s_j;x)\geq aU_j-b.
\label{eq:app-calibration}
\end{equation}

\paragraph{Proposition 3 (thresholded downstream bound).}
Under Assumption~1 and Equation~\ref{eq:app-uniform-error}, selecting a
candidate only when $\widehat U_j>t$ guarantees
\begin{equation}
G(s_j;x)>a(t-\varepsilon)-b.
\label{eq:app-return-bound}
\end{equation}
In particular, $t\geq\varepsilon+b/a$ is sufficient for a nonnegative
downstream-gain guarantee.

\emph{Proof.}
The estimation bound implies $U_j>t-\varepsilon$. Substitution into
Equation~\ref{eq:app-calibration} yields
$G(s_j;x)>a(t-\varepsilon)-b$. \hfill$\square$

The bound exposes a direct tradeoff between the threshold and the guaranteed
downstream margin. Our experiments use $t=0$, while the same gate supports a
tunable positive margin when desired.

\subsection{Interaction and Computation Complexity}

With $M$ candidates and $K$ evaluation episodes per condition, direct rollout
utility estimation requires $(M+1)K$ additional environment trajectories per query when
the skill-free condition is shared. \method launches no candidate-specific
environment trajectories. It reuses $N$ Scouts already collected for training
and requires $MN$ candidate-conditioned trajectory scores plus $N$ cached
skill-free scores, giving $\mathcal{O}(MN)$ additional model scoring work.
After selection, the $L$ post-selection trajectories are ordinary Executor
training data rather than candidate-evaluation rollouts. With the default
$N=M=L=4$, each update uses eight environment trajectories in the Executor group
and twenty reusable trajectory-context scores for utility computation.
\clearpage
\raggedbottom
\section{Skill-Guided Execution Examples}
\label{app:prompts-cases}

The following pages provide the skill-generation prompt, the complete selected
skill, and one skill-guided trajectory per benchmark. The ALFWorld and
ScienceWorld prompts are abridged for space, while preserving their evidence
rules, input structure, and required output format.

\subsection{Qualitative Case of Skill-Free Strategy Acquisition}
\label{app:skill-free-acquisition-case}
\label{app:internalization-case}

\begin{figure}[H]
    \centering
    \includegraphics[width=0.5\textwidth]{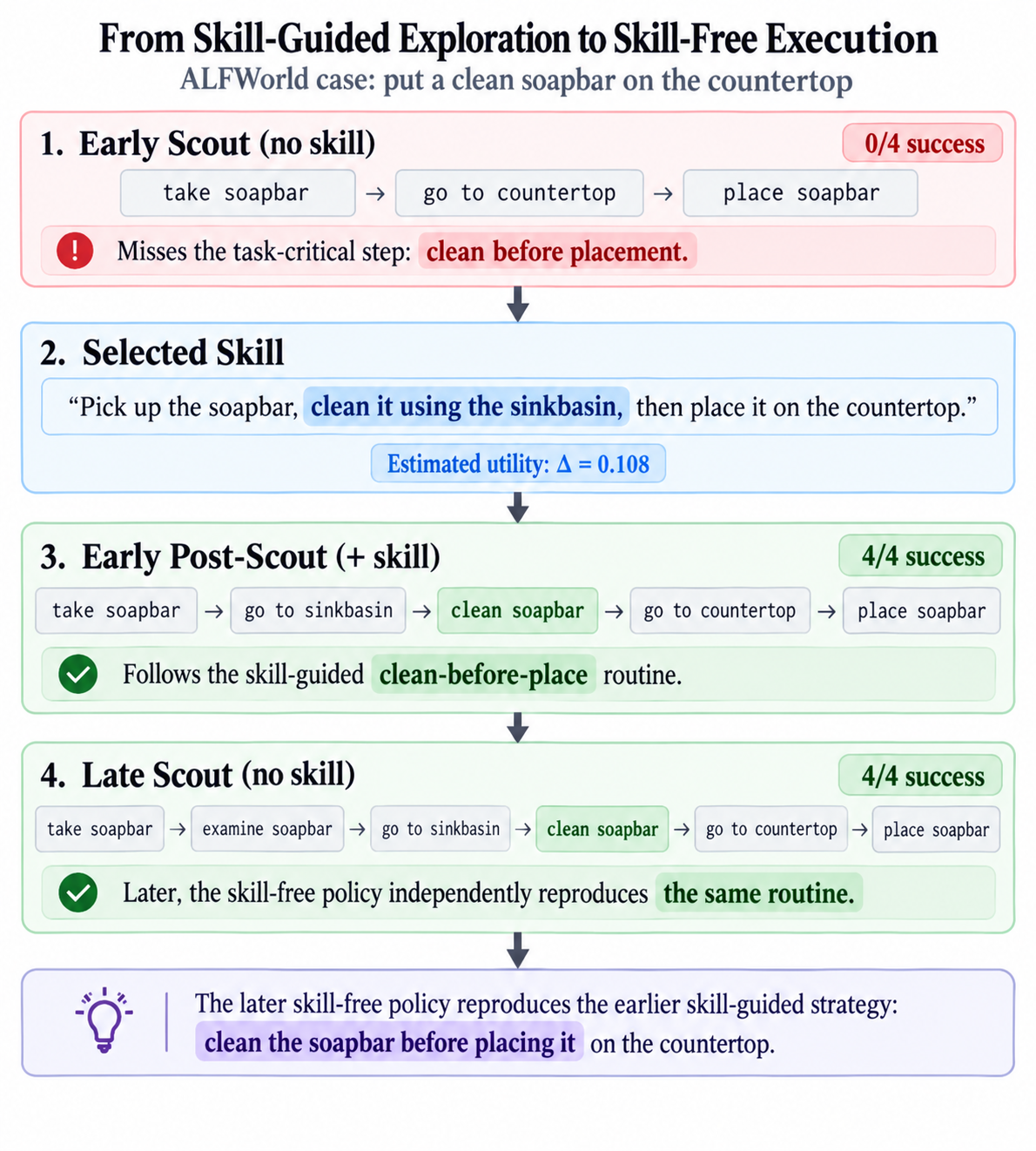}
    \caption{\textbf{From skill-guided exploration to skill-free execution.}
    An ALFWorld example in which a selected skill introduces the missing
    clean-before-place strategy, which is later reproduced without skill
    input.}
    \label{fig:skill-free-acquisition-case}
\end{figure}

Figure~\ref{fig:skill-free-acquisition-case} shows that the early skill-free Scouts
omit the required cleaning step, whereas the selected positive-utility skill
enables all Post-Scouts to complete the task. At a later checkpoint, the
skill-free policy independently reproduces the same clean-before-place routine,
illustrating that the learned policy later executes the strategy without
external skill input.

\definecolor{caseblue}{HTML}{2F65A7}
\definecolor{casegray}{HTML}{536171}
\newcommand{\caseparskip}{3pt}
\newtcolorbox{casebox}[2]{
  colback=white, colframe=#1!65!white,
  colbacktitle=#1!9!white, coltitle=black,
  title={#2}, fonttitle=\bfseries\small,
  fontupper=\fontsize{9}{10.7}\selectfont,
  boxrule=0.4pt, arc=1.8pt,
  left=7pt, right=7pt, top=5pt, bottom=5pt,
  toptitle=3pt, bottomtitle=3pt,
  before skip=5pt, after skip=5pt,
  before upper={\setlength{\parskip}{\caseparskip}\setlength{\parindent}{0pt}\raggedright}
}
\newtcolorbox{caseprompt}[1]{
  breakable,
  colback=white, colframe=casegray!65!white,
  colbacktitle=casegray!9!white, coltitle=black,
  title={#1}, title after break={#1 (continued)},
  fonttitle=\bfseries\small,
  fontupper=\fontsize{8.5}{10.1}\selectfont,
  boxrule=0.4pt, arc=1.8pt,
  left=7pt, right=7pt, top=5pt, bottom=5pt,
  toptitle=3pt, bottomtitle=3pt,
  before skip=5pt, after skip=5pt,
  before upper={\setlength{\parskip}{1pt}\setlength{\parindent}{0pt}\raggedright}
}
\newtcolorbox{caseskill}[1]{
  breakable,
  colback=white, colframe=caseblue!65!white,
  colbacktitle=caseblue!9!white, coltitle=black,
  title={#1}, title after break={#1 (continued)},
  fonttitle=\bfseries\small,
  fontupper=\fontsize{9}{10.7}\selectfont,
  boxrule=0.4pt, arc=1.8pt,
  left=7pt, right=7pt, top=5pt, bottom=5pt,
  toptitle=3pt, bottomtitle=3pt,
  before skip=5pt, after skip=5pt,
  before upper={\setlength{\parskip}{1pt}\setlength{\parindent}{0pt}\raggedright}
}
\newcommand{\caseomit}[1]{{\color{casegray}\itshape[#1]}}
\newcommand{\caseturn}[2]{\par\textbf{Turn #1.}\ \texttt{#2}\par}
\newcommand{\caseobs}[1]{\textit{Observation:} #1\par}
\newcommand{\casefield}[2]{\par\textbf{#1}\ #2\par}

\subsection{ALFWorld: Clean Before Placement}
\label{app:case-alfworld}
\noindent\textbf{Task:} ``put a clean soapbar in countertop.''
The selected skill inserts cleaning between retrieval and placement.

\begin{caseprompt}{1. Skill generation prompt (abridged template)}
\textit{Repeated constraints and the full teaching example are omitted for space.}
\par
\textbf{System message}\par
You extract reusable decision rules for an ALFWorld policy from four
independent attempts at the same task. Write one concise skill that guides a
new attempt using its current observations. Improve executable decisions
rather than reproducing one completed trajectory.
\par
\textbf{Evidence and comparison}
\par
- The task defines the required object, quantity, state, and destination.
Treat observations, inventory, admissible actions, and environment feedback as
evidence; treat the agent's reasoning as an unverified belief.
\par
- SUCCESS/FAILURE labels describe the final outcome only. Compare local
decisions against their visible preconditions and feedback, preserving valid
steps even when the episode fails.
\par
- Ground corrections in rejected actions, confirmed state changes, or missing
observations. For all-failure groups, retain supported progress and recovery
rules without presenting an unobserved complete solution as verified.
\par
\textbf{Abstraction and executability}
\par
- Express the target object, source, destination, required state, and appliance
as semantic roles, then bind them to observed identifiers during execution.
Track the same object instance across retrieval, transformation, and placement.
\par
- Do not copy numbered objects, receptacles, fixed routes, or room layouts.
Use only action forms supported by the interface, and verify explicit state
changes such as cleaning, cooling, or heating before placement.
\par
- Distinguish visiting a location from inspecting its contents. A failed search
does not prove absence, and a rejected action should not be repeated without
checking its command or preconditions.
\par
\textbf{Output discipline}
\par
Return exactly one concise card enclosed by
\texttt{\textless skill\textgreater} and
\texttt{\textless/skill\textgreater}. Each rule should connect an observable condition to an
action, verification check, or recovery step. Do not mention trajectories,
outcome labels, rewards, demonstrations, or this prompt in the card.
\medskip\par
\textbf{User message}\par
\textbf{Abridged teaching example---all attempts fail}
\par
The attempts show three recurring local errors: treating a closed container as
searched, using an identifier inconsistent with the observation, and repeating
placement while still at the source. The example skill therefore recommends
inspecting hidden contents, binding commands to observed identifiers, reading
rejection feedback, and correcting the relevant precondition before retrying.
\par
\textbf{Real task to process}
\par
Task: \texttt{\{task\}}
\par
Four independent skill-free attempts with environment outcomes:
\par
\texttt{\textless attempt\_records\textgreater}
\par
\texttt{\{trajectories\}}
\par
\texttt{\textless/attempt\_records\textgreater}
\par
Write one coherent skill grounded in these records. Keep it within 400 English
words and use exactly the following headings:
\par
\textbf{\# Skill: \texttt{\textless short-kebab-case-name\textgreater}}
\par
\textbf{\#\# When to apply}
\par
\textbf{\#\# Success-preserving strategy}
\par
\textbf{\#\# Failure patterns to suppress}
\par
\textbf{\#\# Procedure}
\par
\textbf{\#\# Decision rules}
\par
\textbf{\#\# Verification and recovery}
\par
\textbf{\#\# Stop conditions}
\par
Return only \texttt{\textless skill\textgreater...\textless/skill\textgreater}
with the headings above.
\medskip\par
\textbf{Assistant generation prefix}\par
\texttt{\textless think\textgreater}
\par
\texttt{\textless/think\textgreater}

\end{caseprompt}

\begin{caseskill}{2. Selected skill (complete card)}
\textbf{\# Skill: locate-and-clean-soapbar-before-placement}
\par
\textbf{\#\# When to apply}\enspace Use when the task requires locating a soapbar, cleaning it, and placing it on a countertop. The soapbar must be identified, retrieved, and cleaned before placement.
\par
\textbf{\#\# Success-preserving strategy}\enspace Preserve confirmed local progress: keep track of locations inspected, objects found, and actions taken. Distinguish between observed states and unverified assumptions.
\par
\textbf{\#\# Failure patterns to suppress}\enspace Do not assume the soapbar is absent after searching common locations. Do not attempt placement without confirming the soapbar is clean. Avoid repeating the same action without correcting its preconditions.
\par
\textbf{\#\# Procedure}\enspace 1. Inspect common locations for the soapbar, including sinkbasin, cabinets, drawers, and shelves.
\par
2. Once found, pick up the soapbar and clean it using the sinkbasin.
\par
3. After cleaning, move to the designated countertop and place the soapbar there.
\par
\textbf{\#\# Decision rules}\enspace - If the soapbar is not found in common locations, expand the search to less obvious places like the garbagecan.
\par
- If the soapbar is found but not clean, use the sinkbasin to clean it before attempting placement.
\par
- If placement is rejected, verify that the soapbar is clean and located at the correct destination.
\par
\textbf{\#\# Verification and recovery}\enspace Check the feedback after each action to confirm state changes. If the soapbar is not clean, retry the cleaning step. If placement fails, recheck the soapbar's state and location.
\par
\textbf{\#\# Stop conditions}\enspace When the soapbar is clean and placed on the countertop, the task is complete. Budget expiry or incomplete search does not prove absence; continue searching until the soapbar is found and placed.

\end{caseskill}

\begin{casebox}{caseblue}{3. Skill-guided trajectory (one Post-Scout; key turns)}
\caseomit{Turns 0--19: search omitted. Original turn indices are zero-based.}
\caseturn{20}{take soapbar 1 from toilet 1}
\caseobs{You pick up the soapbar 1 from the toilet 1.}
\caseturn{21}{go to sinkbasin 1}
\caseobs{You arrive at sinkbasin 1. On the sinkbasin 1, you see nothing.}
\caseturn{22}{clean soapbar 1 with sinkbasin 1}
\caseobs{You clean the soapbar 1 using the sinkbasin 1.}
\caseturn{23}{go to countertop 1}
\caseobs{You arrive at countertop 1. On the countertop 1, you see a toiletpaper 1.}
\caseturn{24}{move soapbar 1 to countertop 1}
\caseomit{Terminal metadata: won=true, done=true. No subsequent textual observation was saved.}
\end{casebox}

\clearpage
\subsection{WebShop: Verify the Required Option Before Purchase}
\label{app:case-webshop}
\noindent\textbf{Task:} ``i need a button down shirt with a long sleeve for a
evereday wear medium size with v neck, and price lower than 50.00 dollars.''
This example uses the anchor-based variant and its actual successful Scout
as Generator evidence.

\begin{caseprompt}{1. Skill generation prompt (complete template)}
\textbf{System message}\par
You are the Skill Extractor in a reinforcement-learning WebShop agent. Given one trajectory that the environment confirms succeeded, extract exactly one reusable policy skill grounded only in the supplied evidence. Success does not imply that every action in the trajectory was useful or optimal.
\medskip\par
\textbf{User message}\par
You are the Skill Extractor in a reinforcement-learning WebShop agent. Extract the single most reusable decision mechanism supported by the private successful evidence below.
\par
Private task: \{task\_description\} Private trajectory: \{trajectory\}
\par
STRICT SERIALIZATION CONTRACT — violation makes the entire output invalid:
\par
1. The first character of your response must be `\{` and the last character must be `\}`.
\par
2. Emit exactly one valid JSON object with exactly one key named `skill`.
\par
3. Never emit backticks, a code fence, analysis, preface, or trailing text.
\par
4. Encode newlines inside the JSON string normally; ensure `json.loads` can parse the complete response.
\par
The `skill` value is a 50-150 word Markdown card with headings in this exact order: `\# Skill:`, `\#\# When to Use`, `\#\# Inputs to Identify`, `\#\# Workflow`, `\#\# Decision Rules`. Use no more than six complete sentences; list fragments are allowed.
\par
Choose exactly one mechanism name: `query-reformulation-after-weak-retrieval`, `explicit-contradiction-screening`, `page-evidence-constraint-verification`, `required-option-verification-before-purchase`, `compatibility-evidence-verification`, or `budget-and-constraint-gating`.
\par
Use only these semantic role labels for request-specific concepts: product type, hard constraint, contradictory evidence, required option, compatibility requirement, page evidence, budget, candidate, search result. Keep them verbatim and never add a concrete example or subtype. Invalid: `device compatibility`, `device family`, `color or size`, `fragrance`, `power rating`, `running use`. Valid replacements: `compatibility requirement`, `required option`, or `hard constraint`. Do not output any product/category name, brand, identifier, query phrase, number, currency amount, size, color, material, audience, device, power rating, scent, skin property, sport/use case, or requested attribute. Do not mention private evidence, task, source, trajectory, example, outcome, reward, or success.
\par
Describe only the selected mechanism: its observable trigger, ordered action, and verification or stopping condition. Do not summarize the full episode, invent controls or facts, or claim universal optimality. Silently verify that the card would remain valid for a different product category without editing.
\par
Your response must begin exactly like this (with the selected name replacing the placeholder): \{"skill":"\# Skill: mechanism-name\textbackslash{}n
\medskip\par\textbf{Assistant generation prefix}\par
\texttt{<think>}
\par
\texttt{</think>}

\end{caseprompt}

\begin{caseskill}{2. Selected skill (complete card)}
\textbf{\# Skill: required-option-verification-before-purchase}
\par
\textbf{\#\# When to Use}\enspace When a product meets all hard constraints but lacks a required option.
\par
\textbf{\#\# Inputs to Identify}\enspace A product that satisfies price, type, and compatibility requirements but is missing a required option.
\par
\textbf{\#\# Workflow}\enspace Check if the required option is available in the specified size or color.
\par
\textbf{\#\# Decision Rules}\enspace If the required option is available, proceed with purchase; otherwise, backtrack to refine search.

\end{caseskill}

\begin{casebox}{caseblue}{3. Skill-guided trajectory (one Post-Scout; all four actions)}
\caseomit{Reasoning omitted; page observations excerpted. Turn indices are zero-based.}
\caseturn{0}{search[button down shirt long sleeve v neck medium]}
\caseobs{\ldots\ B09KR6LR84 [SEP] Shirt for Women Button Down Tops Colorful
Blouses Casual V Neck Sexy Long Sleeve T-Shirts [SEP] \$25.99\ \ldots}
\caseturn{1}{click[b09kr6lr84]}
\caseobs{\ldots\ size [SEP] small [SEP] medium [SEP] large [SEP] xx-large
[SEP] 3x-large\ \ldots\ Price: \$25.99\ \ldots}
\caseturn{2}{click[medium]}
\caseobs{\ldots\ size [SEP] small [SEP] medium [SEP] large [SEP] xx-large
[SEP] 3x-large\ \ldots\ Price: \$25.99\ \ldots}
\caseturn{3}{click[buy now]}
\caseomit{Terminal metadata: won=true, done=true, task\_score=1.
No subsequent textual observation was saved.}
\end{casebox}

\clearpage
\subsection{ScienceWorld: Complete the Circuit Before Classifying}
\label{app:case-scienceworld}
\noindent\textbf{Task (excerpt):} ``Your task is to determine if metal fork is
electrically conductive.'' The task assigns conductive objects to the red
box and nonconductive objects to the green box.

\begin{caseprompt}{1. Skill generation prompt (abridged template)}
\textit{Repeated constraints and the full teaching example are omitted for space.}
\par
\textbf{System message}\par
You extract reusable ScienceWorld laboratory skills from four attempts at the
same task. Learn observable decision rules from successful and unsuccessful
attempts, and use only the supplied evidence. A failed or budget-limited
attempt may still contain valid partial progress.
\par
Never invent instruments, commands, object properties, causal effects, or
experimental results. In particular, an absent signal is not a negative
material test until the observed setup is complete and verified. Missing
equipment, invalid commands, incomplete circuits, and action limits must not be
converted into scientific conclusions.
\par
Abstract exact object instances, rooms, colors, and destinations into roles
read from the current task. Preserve scientific properties and action forms
only when they are supported by the observed attempts.
\par
Return one evidence-grounded card of at most 400 English words, enclosed by
\texttt{\textless skill\textgreater} and
\texttt{\textless/skill\textgreater}.
\medskip\par
\textbf{User message}\par
Task: \texttt{\{task\}}
\par
Four observed attempts with environment-confirmed outcomes:
\par
\texttt{\{trajectories\}}
\par
Produce a reusable procedure rather than a transcript tied to one variation.
Preserve verified progress, identify supported failure-producing decisions,
and bind commands to currently observable objects. Distinguish measured facts
from hypotheses, and check equipment, material state, and prerequisites before
interpreting an unsuccessful experiment.
\par
\textbf{Abridged teaching example---verify before classifying}
\par
For a conductivity task with no clear signal, first inspect the available
equipment and connections. Correct only problems supported by feedback, then
retest and classify the material only when the task's observable criterion is
met. If the setup remains incomplete, retain uncertainty rather than assigning
an unsupported material property.
\par
Use exactly the following structure:
\par
\texttt{\textless skill\textgreater}
\par
\textbf{\# Skill: short reusable mechanism name}
\par
\textbf{\#\# When to apply}
\par
\textbf{\#\# Success-preserving strategy}
\par
\textbf{\#\# Failure patterns to suppress}
\par
\textbf{\#\# Procedure}
\par
\textbf{\#\# Decision rules}
\par
\textbf{\#\# Verification and recovery}
\par
\textbf{\#\# Stop conditions}
\par
\texttt{\textless/skill\textgreater}
\par
Output only the card. Do not mention rewards, trajectories, demonstrations, or
fixed private box colors and room names.
\medskip\par
\textbf{Assistant generation prefix}\par
\texttt{\textless think\textgreater}
\par
\texttt{\textless/think\textgreater}

\end{caseprompt}

\begin{caseskill}{2. Selected skill (complete card)}
\textbf{\# Skill: Test Conductivity Using Available Setup}
\par
\textbf{\#\# When to apply}\enspace A material is suspected to be electrically conductive, and a test setup involving a power source and a light bulb is available.
\par
\textbf{\#\# Success-preserving strategy}\enspace Maintain the material in the inventory, confirm the presence of a power source and light bulb, and ensure the setup is connected correctly. Use the task's specified destination for classification.
\par
\textbf{\#\# Failure patterns to suppress}\enspace Using unverified equipment, attempting to use non-existent tools like a conductivity tester, or connecting the material without a complete circuit.
\par
\textbf{\#\# Procedure}\enspace 1. Focus on the metal fork to examine its properties.
\par
2. Pick up the metal fork to place it in the inventory.
\par
3. Check for the presence of a power source and light bulb in the environment.
\par
4. Connect the metal fork to the power source and light bulb to form a circuit.
\par
5. Observe if the light bulb illuminates to determine conductivity.
\par
6. If the light bulb lights up, place the metal fork in the destination specified by the current task.
\par
\textbf{\#\# Decision rules}\enspace If the light bulb does not illuminate, check the connections and ensure the circuit is complete. If the power source or light bulb is not available, do not proceed with the test. If the connection command is rejected, verify the presence of the required objects before retrying.
\par
\textbf{\#\# Verification and recovery}\enspace Confirm that the light bulb lights up as a clear signal of conductivity. If the light bulb does not respond, check the setup for errors without assuming a material property. If the setup is confirmed but the signal is absent, retain uncertainty and avoid classification.
\par
\textbf{\#\# Stop conditions}\enspace Stop when the light bulb illuminates, confirming conductivity, or when the setup is confirmed to be incomplete and no further actions can be taken.

\end{caseskill}

\begin{casebox}{caseblue}{3. Skill-guided trajectory (one Post-Scout; key turns)}
\caseomit{Turns 0--10: search, pickup, and navigation omitted. Zero-based indices.}
\caseturn{11}{connect metal fork to battery}
\caseobs{Connections must specify specific points of contact on polarized
electrical components (e.g. anode, cathode)}
\caseturn{12}{connect metal fork to anode in battery}
\caseobs{terminal 1 on metal fork is now connected to anode on battery}
\caseturn{13}{connect metal fork to blue light bulb cathode}
\caseobs{terminal 2 on metal fork is now connected to cathode on blue light bulb}
\caseturn{14}{look at blue light bulb}
\caseobs{a blue light bulb, which is off. its anode is connected to: nothing.
its cathode is connected to: a terminal 2 on metal fork.}
\caseomit{Turns 15--16: inspections omitted.}
\caseturn{17}{connect battery cathode to blue light bulb anode}
\caseobs{cathode on battery is now connected to anode on blue light bulb}
\caseturn{18}{look at blue light bulb}
\caseobs{a blue light bulb, which is on. its anode is connected to: a cathode
on battery. its cathode is connected to: a terminal 2 on metal fork.}
\caseturn{19}{move metal fork to red box}
\caseomit{Terminal metadata: won=true, done=true, task\_score=100.
No subsequent textual observation was saved.}
\end{casebox}

\clearpage
\begingroup
\renewcommand{\caseparskip}{1pt}
\subsection{\texorpdfstring{$\tau^2$-Bench}{Tau2-Bench}: Confirm an Exchange Before Executing It}
\label{app:case-tau}
\noindent\textbf{Task (summary):} Exchange a delivered action camera for an
available 4K, waterproof, black variant of the same product.

\begin{caseprompt}{1. Skill generation prompt (complete template)}
\textbf{System message}\par
You are a strategy extractor for a tool-using customer-service agent. You receive the exact agent-visible context and four independent attempts for one task. Infer one reusable strategy that preserves mechanisms shared by good attempts and suppresses mechanisms associated with bad attempts. The domain policy and tool schemas always override the strategy. Never reveal or copy task-specific entities. Return only the requested Markdown skill card and nothing else.
\medskip\par
\textbf{User message}\par
DOMAIN: retail
\par
\textbf{RETAIL FOCUS}
\par
- Distinguish order/item lookup from return, exchange, refund, address, payment, and cancellation writes.
\par
- Verify identity, order status, item eligibility, delivery state, quantities, refund destination, replacement availability, and confirmation boundaries before irreversible actions.
\par
- Handle partial orders and multi-item requests explicitly; stop or escalate when policy forbids an action.
\par
- Keep order IDs, product IDs, names, addresses, dates, amounts, and payment IDs abstract.
\par
- Do not name a product category or repeat a user's subjective product preference; say `item` and `requested variant` so the mechanism transfers across products.
\par
\texttt{<AGENT\_VISIBLE\_CONTEXT>}
\par
\texttt{\{visible\_context\}}
\par
\texttt{</AGENT\_VISIBLE\_CONTEXT>}
\par
\texttt{<ATTEMPT index=0 outcome=\{outcomes[0]\}>}
\par
\texttt{\{attempts[0]\}}
\par
\texttt{</ATTEMPT>}
\par
\texttt{<ATTEMPT index=1 outcome=\{outcomes[1]\}>}
\par
\texttt{\{attempts[1]\}}
\par
\texttt{</ATTEMPT>}
\par
\texttt{<ATTEMPT index=2 outcome=\{outcomes[2]\}>}
\par
\texttt{\{attempts[2]\}}
\par
\texttt{</ATTEMPT>}
\par
\texttt{<ATTEMPT index=3 outcome=\{outcomes[3]\}>}
\par
\texttt{\{attempts[3]\}}
\par
\texttt{</ATTEMPT>}
\par
\textbf{OUTPUT CONTRACT}
\par
Return the skill card directly as Markdown, beginning with `\# Skill:` and containing these sections in order: `\#\# When to apply`, `\#\# Policy and tool checks`, `\#\# Procedure`, `\#\# Decision rules`, `\#\# Verification and recovery`, and `\#\# Stop or escalation conditions`. Do not wrap it in JSON or a code fence.
\par
The skill should be 120-240 English words, with no more than two concise sentences or bullets per section. It must be operational rather than generic and reusable on a different query in the same domain. Stable tool names and argument roles are allowed. Do not copy any customer name, account/order/reservation/payment identifier, email, phone number, address, route, flight number, date, price, exact task value, or hidden evaluator language. Do not compare or refer to the supplied examples, their outcome labels, trajectories, rewards, Delta, demonstrations, or grading. Do not solve the concrete query. Extract one mechanism, not a broad customer-service handbook.
\par
\textbf{FINAL RETAIL CHECK}
\par
Recheck that identifiers and product preferences are abstract, every mutation has its required status/eligibility/confirmation checks, and the card transfers to another item or order.

\end{caseprompt}

\begin{caseskill}{2. Selected skill (complete card)}
\textbf{\# Skill: Exchange item for same product type in delivered order}
\par
\textbf{\#\# When to apply}\enspace Use to exchange an item from a delivered order for a different variant of the same product type.
\par
\textbf{\#\# Policy and tool checks}\enspace - Verify order is delivered
\par
- Confirm exchangeable variant is available
\par
- Request refund method (original payment or gift card)
\par
- Ensure only one exchange per order
\par
\textbf{\#\# Procedure}\enspace 1. Fetch user and order details
\par
2. Identify target item and eligible variants
\par
3. Present options and price difference
\par
4. Confirm exchange details before proceeding
\par
\textbf{\#\# Decision rules}\enspace - Reject exchanges to different product types
\par
- Prevent multiple exchanges per order
\par
- Escalate if payment method incompatibility detected
\par
\textbf{\#\# Verification and recovery}\enspace - Confirm refund method matches eligibility
\par
- Store exchange details in order history
\par
- Provide return instructions via email
\par
\textbf{\#\# Stop or escalation conditions}\enspace Transfer if request exceeds exchange scope (e.g., full order modifications).

\end{caseskill}

\begin{casebox}{caseblue}{3. Skill-guided trajectory (one Post-Scout; key turns)}
\caseomit{Identity lookup and other dialogue omitted. Assistant turns are 1-based;
tool results are field excerpts.}
\caseturn{4}{\{"name": "get\_order\_details", "arguments": \{"order\_id": "\#W6759506"\}\}}
\caseobs{\texttt{\ldots\ "status": "delivered"\ \ldots}}
\caseturn{5}{\{"name": "get\_product\_details", "arguments": \{"product\_id": "3377618313"\}\}}
\caseobs{\texttt{\ldots\ "6700049080": \{"item\_id": "6700049080",
"options": \{"resolution": "4K", "waterproof": "yes", "color": "black"\},
"available": true, "price": 466.75\}\ \ldots}}
\par\textbf{Turn 7. Agent (excerpt):}\par
\textbf{Price difference:} -\$36.76 (you'll receive a refund)\par
\textbf{Refund method:} Gift Card (current balance: \$200.00)\par
\ldots\ Do you confirm this exchange? Please reply \textbf{yes} or \textbf{no}.
\par\textit{User:} Yes, I confirm the exchange.
\caseturn{8}{\{"name": "exchange\_delivered\_order\_items", "arguments":
\{"new\_item\_ids": ["6700049080"], "payment\_method\_id": "gift\_card\_4444444",
"item\_ids": ["5925362855"], "order\_id": "\#W6759506"\}\}}
\caseobs{\texttt{\ldots\ "status": "exchange requested"\ \ldots\
"exchange\_price\_difference": -36.76\ \ldots}}
\caseomit{Recorded outcome: reward=1, database match=true.
This confirms an exchange request, not physical return or refund completion.}
\end{casebox}

\endgroup

\end{document}